\documentclass{article}

\PassOptionsToPackage{numbers,sort&compress}{natbib}
\usepackage[final]{neurips_2021}

\usepackage[table]{xcolor}

\usepackage[utf8]{inputenc} 
\usepackage[T1]{fontenc}    
\usepackage{hyperref}       
\usepackage{url}            
\usepackage{booktabs}       
\usepackage{amsfonts}       
\usepackage{nicefrac}       
\usepackage{microtype}      
\usepackage{xcolor}         
\usepackage{xspace}
\usepackage{graphicx}
\usepackage{subfigure}
\usepackage{multirow}
\usepackage{glossaries}
\usepackage[normalem]{ulem}
\usepackage{titlesec}
\usepackage{floatrow}
\usepackage{wrapfig}

\usepackage{dsfont}
\newfloatcommand{capbtabbox}{table}[][\FBwidth]

\usepackage{caption}

\usepackage{amsmath,amsfonts,bm}

\def\eqref#1{equation~\ref{#1}}

\def\1{\bm{1}}

\def\vtheta{{\bm{\theta}}}
\def\vdelta{{\bm{\delta}}}
\def\va{{\bm{a}}}

\def\vx{{\bm{x}}}

\DeclareMathAlphabet{\mathsfit}{\encodingdefault}{\sfdefault}{m}{sl}
\SetMathAlphabet{\mathsfit}{bold}{\encodingdefault}{\sfdefault}{bx}{n}

\def\sA{{\mathbb{A}}}

\def\sS{{\mathbb{S}}}

\newcommand{\E}{\mathbb{E}}

\DeclareMathOperator*{\argmin}{arg\,min}

\DeclareMathOperator{\sign}{sign}

\DeclareMathOperator*{\maximize}{max}

\DeclareMathOperator{\proj}{proj}

\definecolor{header}{gray}{0.9}
\definecolor{subheader}{rgb}{0.63, 0.79, 0.95}
\newcommand{\Tstrut}{\rule{0pt}{2.6ex}}
\newcommand{\Bstrut}{\rule[-0.9ex]{0pt}{0pt}}
\newcommand{\TBstrut}{\Tstrut\Bstrut}

\newacronym{pgd}{PGD}{Projected Gradient Descent}
\newacronym{bim}{BIM}{Basic Iterative Method}
\newacronym{fgsm}{FGSM}{Fast Gradient Sign Method}
\newacronym{wrn}{\textsc{Wrn}}{Wide ResNet}
\newacronym{sgd}{SGD}{Stochastic Gradient Descent}
\newacronym{ddpm}{DDPM}{Denoising Diffusion Probabilistic Model}
\newacronym{vdvae}{VDVAE}{Very Deep Variational Auto-Encoder}
\newacronym{fid}{FID}{Frechet Inception Distance}
\newacronym{is}{IS}{Inception Score}
\newacronym{gan}{GAN}{Generative Adversarial Network}
\newacronym{vae}{VAE}{Variational AutoEncoder}

\newcommand{\cifar}{\textsc{Cifar-10}\xspace}

\newcommand{\cifarh}{\textsc{Cifar-100}\xspace}
\newcommand{\tinyimages}{\textsc{80M-Ti}\xspace}

\newcommand{\tinyimagenet}{\textsc{TinyImageNet}\xspace}

\newcommand{\svhn}{\textsc{Svhn}\xspace}
\newcommand{\linf}{\ensuremath{\ell_\infty}\xspace}
\newcommand{\ltwo}{\ensuremath{\ell_2}\xspace}
\newcommand{\lp}{\ensuremath{\ell_p}\xspace}
\newcommand{\autoattack}{\textsc{AutoAttack}\xspace}
\newcommand{\autopgd}{\textsc{AutoPgd}\xspace}
\newcommand{\multitargeted}{\textsc{MultiTargeted}\xspace}

\newcommand{\pgd}[1]{\textsc{Pgd}\textsuperscript{$#1$}\xspace}
\newcommand{\xent}{l_{\textrm{ce}}}
\newcommand{\wrn}{\gls*{wrn}\xspace}

\newcommand{\squishlist}{
   \begin{list}{$\bullet$}
    { \setlength{\itemsep}{0pt}      \setlength{\parsep}{3pt}
      \setlength{\topsep}{3pt}       \setlength{\partopsep}{0pt}
      \setlength{\leftmargin}{1.5em} \setlength{\labelwidth}{1em}
      \setlength{\labelsep}{0.5em} } }

\newcommand{\squishend}{
    \end{list}  }

\definecolor{TartOrange}{HTML}{ff2e35}
\definecolor{Orange}{HTML}{ff7825}
\definecolor{Mango}{HTML}{ffc013}
\definecolor{AppleGreen}{HTML}{7cb81b}
\definecolor{Blue}{HTML}{1173b0}
\definecolor{BdazzledBlue}{HTML}{2e58a5}
\definecolor{Purple}{HTML}{5b3590}
\definecolor{Sunglow}{HTML}{FFCA3A}

\title{Data Augmentation Can Improve Robustness}

\author{Sylvestre-Alvise Rebuffi*, Sven Gowal*, Dan Calian, \\
\textbf{Florian Stimberg}, \textbf{Olivia Wiles} {\normalfont and} {\bf Timothy Mann} \\
DeepMind, London\\
\texttt{\{sylvestre,sgowal\}@deepmind.com}
}

\begin{document}

\maketitle

\begin{abstract}
Adversarial training suffers from \emph{robust overfitting}, a phenomenon where the robust test accuracy starts to decrease during training.
In this paper, we focus on reducing robust overfitting by using common data augmentation schemes.
We demonstrate that, contrary to previous findings, when combined with model weight averaging, data augmentation can significantly boost robust accuracy.
Furthermore, we compare various data augmentations techniques and observe that spatial composition techniques work best for adversarial training.
Finally, we evaluate our approach on \cifar against \linf and \ltwo norm-bounded perturbations of size $\epsilon = 8/255$ and $\epsilon = 128/255$, respectively.
We show large absolute improvements of +2.93\% and +2.16\% in robust accuracy compared to previous state-of-the-art methods.
In particular, against \linf norm-bounded perturbations of size $\epsilon = 8/255$, our model reaches 60.07\% robust accuracy without using any external data. 
We also achieve a significant performance boost with this approach while using other architectures and datasets such as \cifarh, \svhn and \tinyimagenet. 
\end{abstract}

\section{Introduction}

Despite their success, neural networks are not intrinsically robust.
In particular, it has been shown that the addition of imperceptible deviations to the input, called adversarial perturbations, can cause neural networks to make incorrect predictions with high confidence \citep{carlini_adversarial_2017,carlini_towards_2017,goodfellow_explaining_2014,kurakin_adversarial_2016,szegedy_intriguing_2013}.
Starting with \citet{szegedy_intriguing_2013}, there has been a lot of work on understanding and generating adversarial perturbations \citep{carlini_towards_2017,athalye_synthesizing_2017}, and on building defenses that are robust to such perturbations \citep{goodfellow_explaining_2014,papernot_distillation_2015,madry_towards_2017,kannan_adversarial_2018}.
Unfortunately, many of the defenses proposed in the literature target failure cases found through specific adversaries, and as such they are easily broken by different adversaries \citep{uesato_adversarial_2018,athalye_obfuscated_2018}.
Among successful defenses are robust optimization techniques like the one by \citet{madry_towards_2017} that learns robust models by finding worst-case adversarial perturbations at each training step before adding them to the training data.
In fact, adversarial training as proposed by \citeauthor{madry_towards_2017} is so effective~\citep{gowal_uncovering_2020} that it is the de facto standard for training adversarially robust neural networks. 
Indeed, since \citet{madry_towards_2017}, various modifications to their original implementation have been proposed \citep{zhang_theoretically_2019,xie_feature_2018,pang_boosting_2020,huang_self-adaptive_2020,rice_overfitting_2020, gowal_uncovering_2020}.

\begin{figure}
\floatbox[{\capbeside\thisfloatsetup{capbesideposition={right,top},capbesidewidth=6cm}}]{figure}[\FBwidth]
{
\caption{Robust accuracy of various models submitted to RobustBench~\cite{croce2020robustbench} against \autoattack~\citep{croce_reliable_2020} on \cifar with \linf perturbations of size $8/255$ displayed in publication order.
Our method builds on \citet{gowal_uncovering_2020} (shown above with 57.20\%) and explores how augmented data can be used to improve robust accuracy by +2.87\% without using any additional external data.\label{fig:history}}%
}
{\includegraphics[width=6cm]{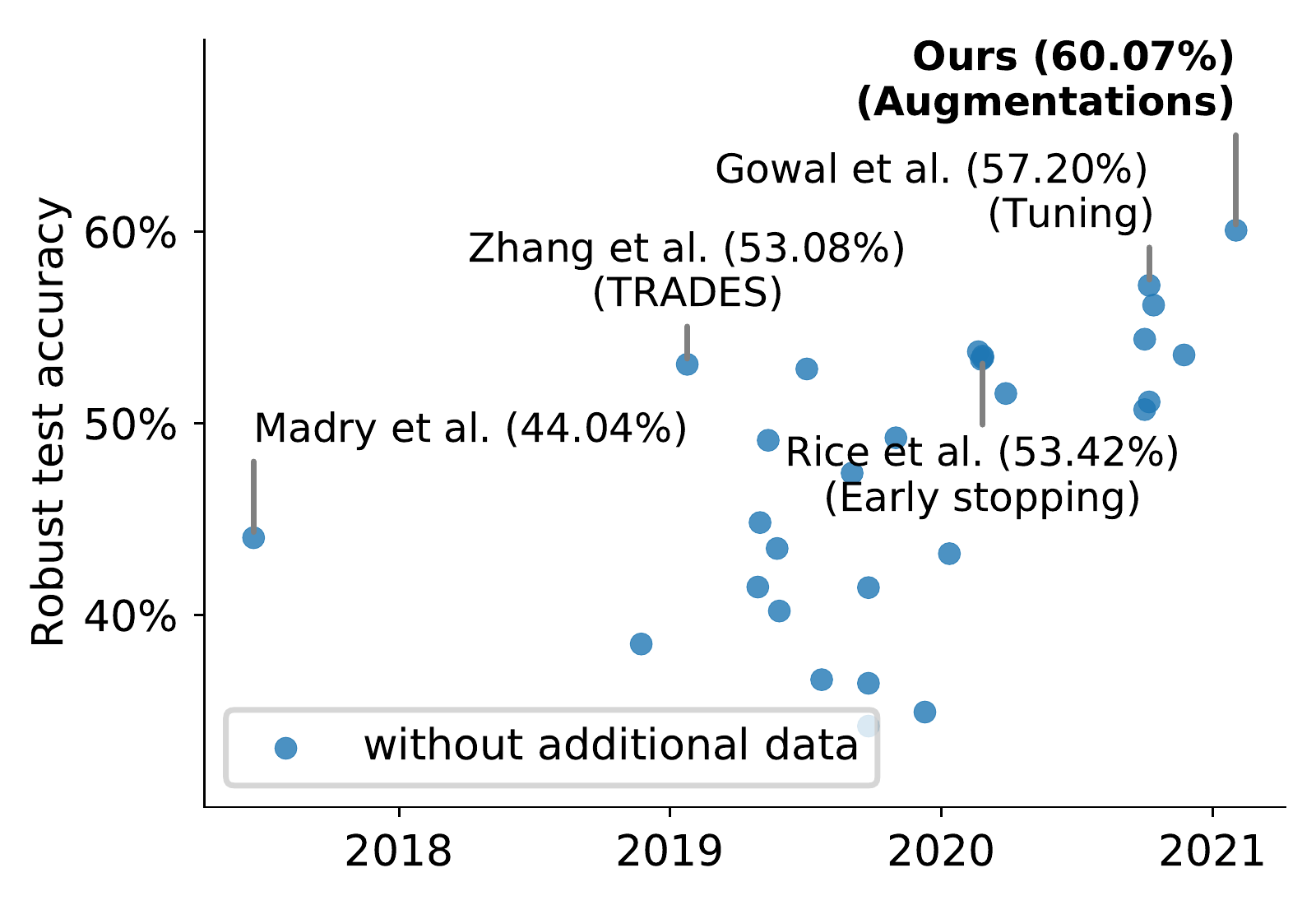}}
\end{figure}

Notably, \citet{hendrycks_using_2019,carmon_unlabeled_2019,uesato_are_2019,zhai_adversarially_2019,najafi_robustness_2019} showed that using additional data improves adversarial robustness, while \citet{rice_overfitting_2020,wu2020adversarial,gowal_uncovering_2020} found that data augmentation techniques did not boost robustness.
This dichotomy motivates this paper.
In particular, we explore whether it is possible to fix the training procedure such that data augmentation becomes useful in the setting without additional data.
By making the observation that model weight averaging (WA)~\citep{izmailov_averaging_2018} helps robust generalization to a wider extent when robust overfitting is minimized, we propose to combine model weight averaging with data augmentation techniques.
Overall, we make the following contributions:
\squishlist
\item We demonstrate that, when combined with model weight averaging, data augmentation techniques such as \emph{Cutout}~\citep{devries2017improved}, \emph{CutMix}~\citep{yun2019cutmix} and \emph{MixUp}~\citep{zhang2017mixup} can improve robustness.
\item To the contrary of \citet{rice_overfitting_2020,wu2020adversarial,gowal_uncovering_2020} which all tried data augmentation techniques without success, we are able to use any of these three aforementioned techniques to obtain new state-of-the-art robust accuracies (see \autoref{fig:history}). We find \emph{CutMix} to be the most effective method by reaching a robust accuracy of 60.07\% on \cifar against \linf perturbations of size $\epsilon = 8/255$ (an improvement of +2.93\% upon the state-of-the-art).
\item We conduct thorough experiments to show that our approach generalizes across architectures, datasets and threat models. We also investigate the trade-off between robust overfitting and underfitting to explain  why \emph{MixUp} performs worse than spatial composition techniques.
\item Finally, we provide empirical evidence that weight averaging exploits data augmentation by ensembling model snapshots which have the same total accuracy but differ at the individual prediction level.
\squishend


\section{Related Work}

\paragraph{Adversarial \lp-norm attacks.}

Since \citet{szegedy_intriguing_2013} observed that neural networks which achieve high accuracy on test data are highly vulnerable to adversarial examples, the art of crafting increasingly sophisticated adversarial examples has received a lot of attention.
\citet{goodfellow_explaining_2014} proposed the \gls{fgsm} which generates adversarial examples with a single normalized gradient step.
It was followed by R+\gls{fgsm} \citep{tramer_ensemble_2017}, which adds a randomization step, and the \gls{bim} \citep{kurakin_adversarial_2016}, which takes multiple smaller gradient steps.

\paragraph{Adversarial training as a defense.}

The adversarial training procedure~\citep{madry_towards_2017} feeds adversarially perturbed examples back into the training data.
It is widely regarded as one of the most successful method to train robust deep neural networks.
It has been augmented in different ways -- with changes in the attack procedure (e.g., by incorporating momentum \cite{dong_boosting_2017}), loss function (e.g., logit pairing \cite{mosbach_logit_2018}) or model architecture (e.g., feature denoising \cite{xie_feature_2018}).
Another notable work by \citet{zhang_theoretically_2019} proposed TRADES, which balances the trade-off between standard and robust accuracy, and achieved state-of-the-art performance against \linf norm-bounded perturbations on \cifar.
More recently, the work from \citet{rice_overfitting_2020} studied \emph{robust overfitting} and demonstrated that improvements similar to TRADES could be obtained more easily using classical adversarial training with early stopping.
This later study revealed that early stopping was competitive with many other regularization techniques and demonstrated that data augmentation schemes beyond the typical \emph{random padding-and-cropping} were ineffective on \cifar.
Finally, \citet{gowal_uncovering_2020} highlighted how different hyper-parameters (such as network size and model weight averaging) affect robustness.
They were able to obtain models that significantly improved upon the state-of-the-art, but lacked a thorough investigation on data augmentation schemes.
Similarly to \citet{rice_overfitting_2020}, they also make the conclusion that data augmentations beyond \emph{random padding-and-cropping} do not improve robustness.

\paragraph{Data augmentation.}

Data augmentation has been shown to improve the generalisation of standard (non-robust) training.
For image classification tasks, random flips, rotations and crops are commonly used~\cite{he2015deep}.
More sophisticated techniques such as \emph{Cutout}~\cite{devries2017improved} (which produces random occlusions), \emph{CutMix}~\cite{yun2019cutmix} (which replaces parts of an image with another) and \emph{MixUp}~\cite{zhang2017mixup} (which linearly interpolates between two images) all demonstrate extremely compelling results.
As such, it is rather surprising that they remain ineffective when training adversarially robust networks~\cite{rice_overfitting_2020,gowal_uncovering_2020,wu2020adversarial}. 
In this work, we revisit these common augmentation techniques in the context of adversarial training.

\section{Preliminaries and hypothesis}
\label{sec:hypothesis}

The rest of this manuscript is organized as follows.
In this section, we provide an overview of adversarial training and introduce the hypothesis that model weight averaging works better when robust overfitting is reduced. In \autoref{sec:heuristics} we discuss that data augmentations can be used to verify this hypothesis.
Finally, we provide thorough experimental results in \autoref{sec:exp}.

\subsection{Adversarial training}

\citet{madry_towards_2017} formulate a saddle point problem to find model parameters $\vtheta$ that minimize the adversarial risk:
\begin{equation}
\argmin_\vtheta \E_{(\vx,y) \sim \mathcal{D}} \left[ \maximize_{\vdelta \in \sS} l(f(\vx + \vdelta; \vtheta), y) \right]
\label{eq:adversarial_risk}
\end{equation}
\noindent where $\mathcal{D}$ is a data distribution over pairs of examples $\vx$ and corresponding labels $y$, $f(\cdot; \vtheta)$ is a model parametrized by $\vtheta$, $l$ is a suitable loss function (such as the $0-1$ loss in the context of classification tasks), and $\sS$ defines the set of allowed perturbations.
For $\ell_p$ norm-bounded perturbations of size $\epsilon$, the adversarial set is defined as $\sS_p = \{ \vdelta ~|~ \| \vdelta \|_p \leq \epsilon \}$.
In the rest of this manuscript, we will use $\epsilon_p$ to denote $\ell_p$ norm-bounded perturbations of size $\epsilon$ (e.g., $\epsilon_\infty = 8/255$).
To solve the inner optimization problem, \citet{madry_towards_2017} use \gls*{pgd}, which replaces the non-differentiable $0-1$ loss $l$ with the cross-entropy loss $\xent$ and computes an adversarial perturbation $\hat{\vdelta} = \vdelta^{(K)}$ in $K$ gradient ascent steps of size $\alpha$ as
\begin{equation}
\vdelta^{(k+1)} \gets \proj_{\sS} \left( \vdelta^{(k)} + \alpha \sign \left(\nabla_{\vdelta^{(k)}} \xent(f(\vx + \vdelta^{(k)}; \vtheta), y) \right)\right)
\label{eq:bim}
\end{equation}
where $\vdelta^{(0)}$ is chosen at random within $\sS$, and where $\proj_{\sA}(\va)$ projects a point $\va$ back onto a set $\sA$, $\proj_{\sA}(\va) = \mathrm{argmin}_{\va' \in \sA} \|\va - \va'\|_2$.
We refer to this inner optimization with $K$ steps as \pgd{K}.

\subsection{Robust overfitting}

To the contrary of standard training, which often shows no \emph{overfitting} in practice~\citep{zhang2017understanding}, adversarial training suffers from \emph{robust overfitting}~\citep{rice_overfitting_2020}.
Robust overfitting is the phenomenon by which robust accuracy on the test set quickly degrades while it continues to rise on the train set (clean accuracy on both sets continues to improve as well).
\citet{rice_overfitting_2020} propose to use early stopping as the main contingency against robust overfitting, and demonstrate that it also allows to train models that are more robust than those trained with other regularization techniques (such as data augmentation or increased \ltwo-regularization).
They observed that some of these other regularization techniques could reduce the impact of overfitting at the cost of producing models that are over-regularized and lack overall robustness and accuracy.
There is one notable exception which is the addition of external data \citep{carmon_unlabeled_2019,uesato_are_2019}.
Figure \ref{fig:no_wa_external_vs_original} shows how the robust accuracy (evaluated on the test set) evolves as training progresses on \cifar against $\epsilon_\infty = 8/255$.
Without external data, robust overfitting is clearly visible and appears shortly after the learning rate is dropped (the learning rate is decayed by 10$\times$ two-thirds through training in a schedule similar to \citep{rice_overfitting_2020} and commonly used since \citep{madry_towards_2017}).
Robust overfitting completely disappears when an additional set of 500K pseudo-labeled images (see \citet{carmon_unlabeled_2019}) is introduced.


\subsection{Model weight averaging}

Model weight averaging (WA)~\citep{izmailov_averaging_2018} can be implemented using an exponential moving average $\vtheta'$ of the model parameters $\vtheta$ with a decay rate $\tau$ (i.e., $\vtheta' \gets \tau \cdot \vtheta' + (1 - \tau) \cdot \vtheta$ at each training step).
During evaluation, the weighted parameters $\vtheta'$ are used instead of the trained parameters $\vtheta$.
\citet{gowal_uncovering_2020,chen2021robust} discovered that model weight averaging can significantly improve robustness on a wide range of models and datasets.
\citet{chen2021robust} argue (similarly to \citep{wu2020adversarial}) that WA leads to a flatter adversarial loss landscape, and thus a smaller robust generalization gap.
\citet{gowal_uncovering_2020} also explain that, in addition to improved robustness, WA reduces sensitivity to early stopping.
While this is true, it is important to note that WA is still prone to robust overfitting.
This is not surprising, since the exponential moving average ``forgets'' older model parameters as training goes on.
Figure \ref{fig:wa_vs_no_wa_robust_overfitting} shows how the robust accuracy evolves as training progresses when using WA.
We observe that, after the change of learning rate, the averaged weights are increasingly affected by overfitting, thus resulting in worse robust accuracy for the averaged model.

\subsection{Hypothesis}

\begin{figure*}[t]
\centering
\subfigure[Adversarial training with and without additional data from \tinyimages (without WA)]{\label{fig:no_wa_external_vs_original}\includegraphics[width=0.3\textwidth]{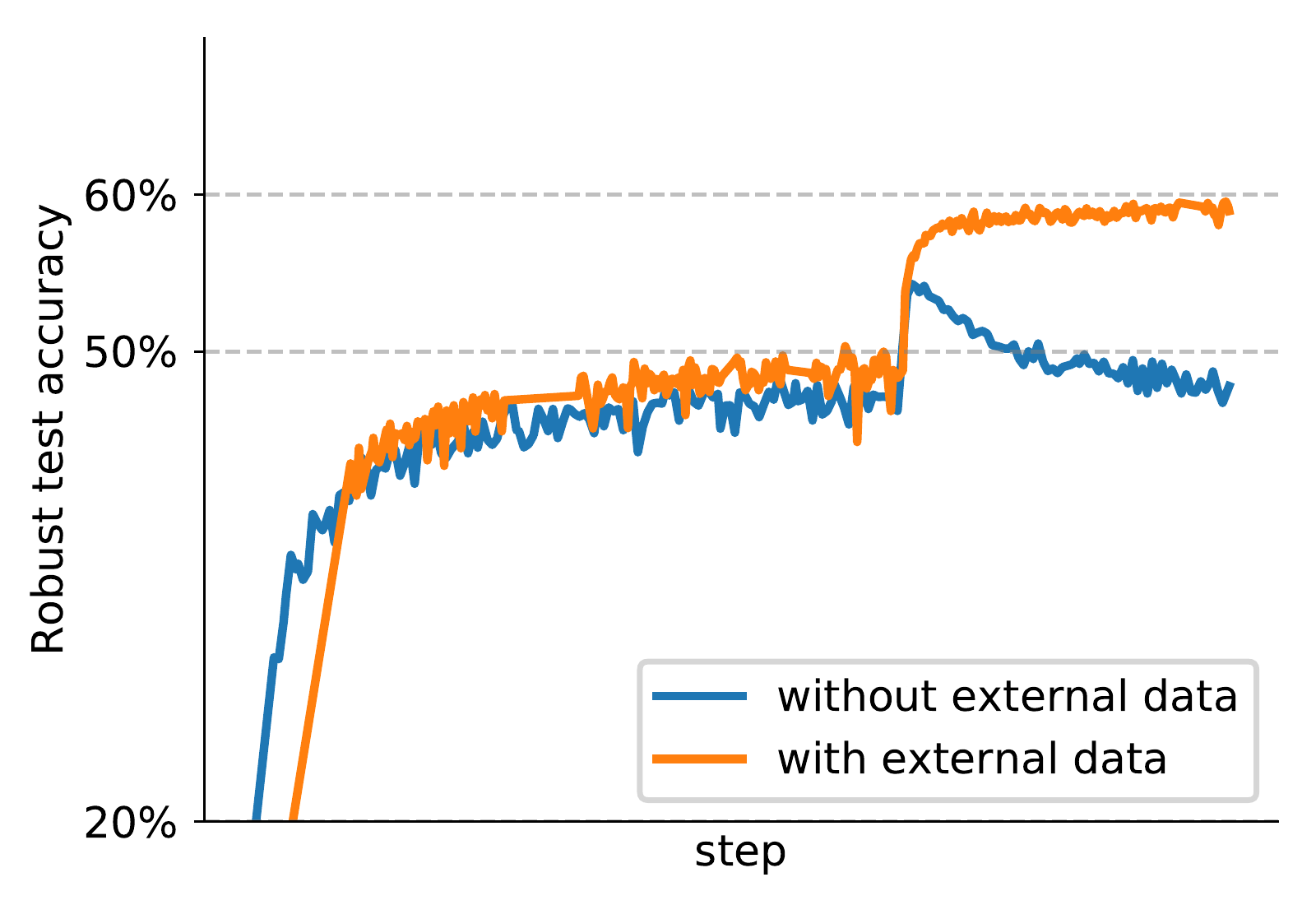}}
\hspace{.5cm}
\subfigure[Effect of WA without external data]{\label{fig:wa_vs_no_wa_robust_overfitting}\includegraphics[width=0.3\textwidth]{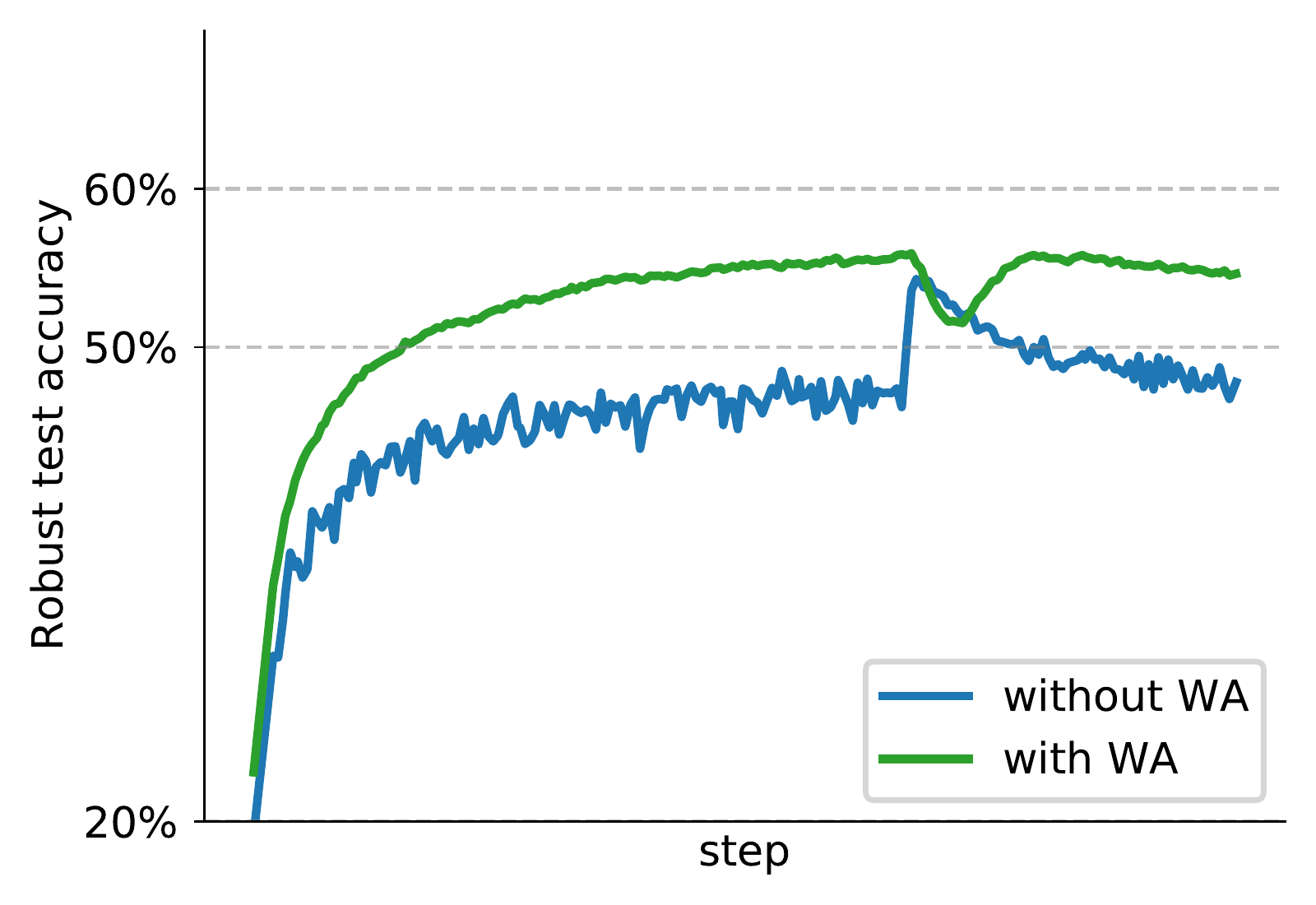}}
\hspace{.5cm}
\subfigure[Effect of WA with external data]{\label{fig:wa_vs_no_wa_external_data}\includegraphics[width=0.3\textwidth]{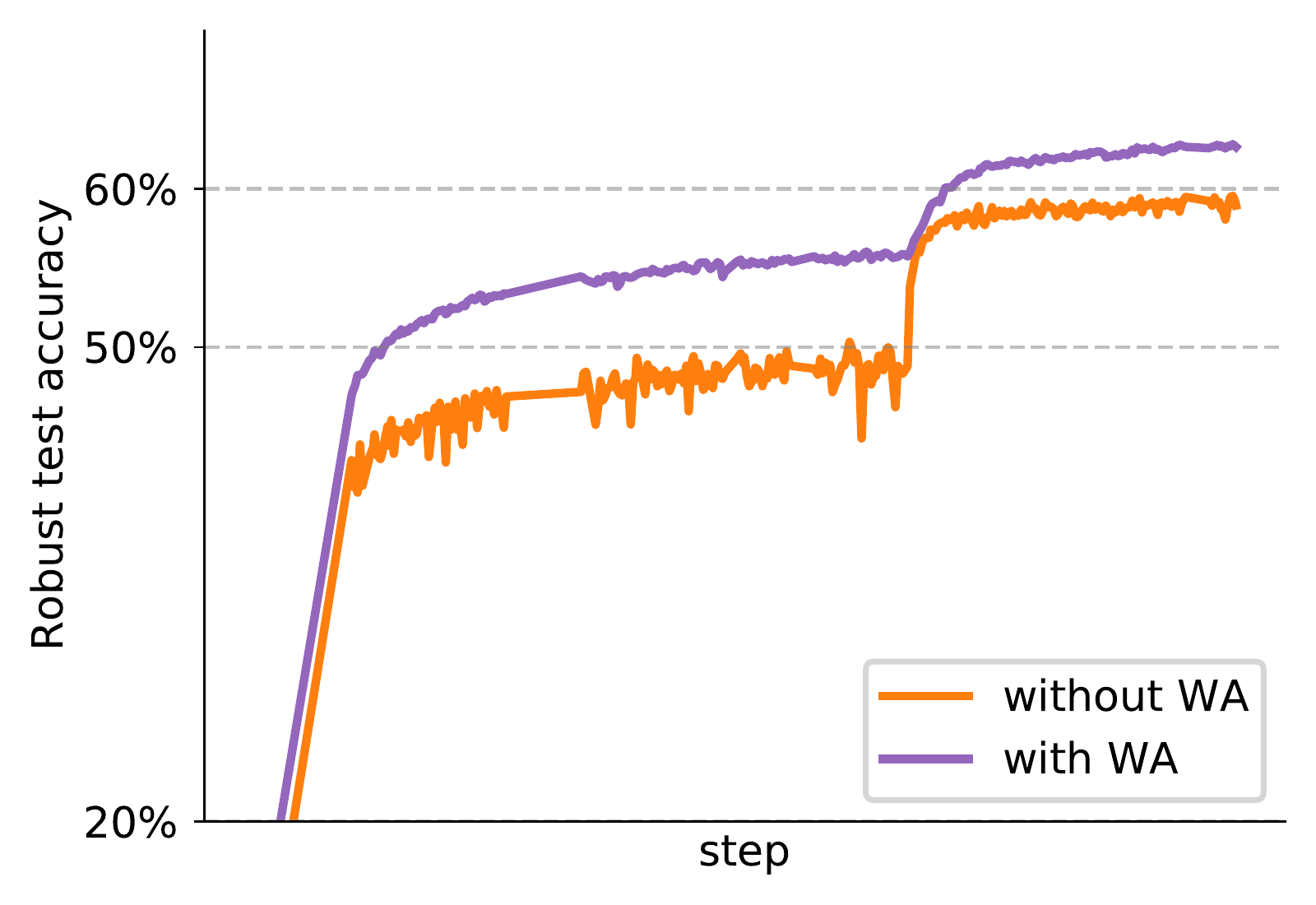}}
\caption{We compare the robust accuracy against $\epsilon_\infty = 8/255$ on \cifar of an adversarially trained \wrn-28-10.
Panel~\subref{fig:no_wa_external_vs_original} shows the impact of using additional external data from \tinyimages~\citep{80m} and illustrates \emph{robust overfitting}.
Panel~\subref{fig:wa_vs_no_wa_robust_overfitting} shows the benefit of \emph{model weight averaging} (WA) despite robust overfitting.
Panel~\subref{fig:wa_vs_no_wa_external_data} shows that WA remains effective and useful even when robust overfitting disappears.
The graphs show the evolution of the robust accuracy as training progresses (against \pgd{40}).
The jump two-thirds through training is due to a drop in learning rate.
\label{fig:robust_overfitting}
}
\end{figure*}

As WA results in flatter, wider solutions compared to the steep decrease in robust accuracy observed for \gls{sgd}~\citep{chen2021robust}, it is natural to ask ourselves whether WA remains useful in cases that do not exhibit robust overfitting.
\autoref{fig:wa_vs_no_wa_external_data} shows how the robust accuracy evolves as training progresses when using WA and additional external data (for which standard SGD does not show signs of overfitting).
We notice that the robust performance in this setting is not only preserved but even boosted when using WA. 
Hence, we formulate the hypothesis that \uline{model weight averaging helps robustness to a greater extent when robust accuracy between model iterations can be maintained}.
This hypothesis is also motivated by the observation that WA acts as a temporal ensemble -- akin to Fast Geometric Ensembling by \citet{garipov2018loss} who show that efficient ensembling can be obtained by aggregating multiple checkpoint parameters at different training times.
As such, to improve robustness, it is important to ensemble a suite of equally strong and diverse models.
Although mildly successful, we note that ensembling has received some attention in the context of adversarial training~\citep{pang2019improving,strauss2017ensemble}.
In particular, \citet{tramer_ensemble_2017,grefenstette2018strength} found that ensembling could reduce the risk of gradient obfuscation caused by locally non-linear loss surfaces.


\section{Data augmentations}
\label{sec:heuristics}

\paragraph{Limiting robust overfitting without external data.}

\begin{figure*}[t]
\centering
\subfigure[Without WA]{\label{fig:augmentations_no_wa}\includegraphics[width=0.35\textwidth]{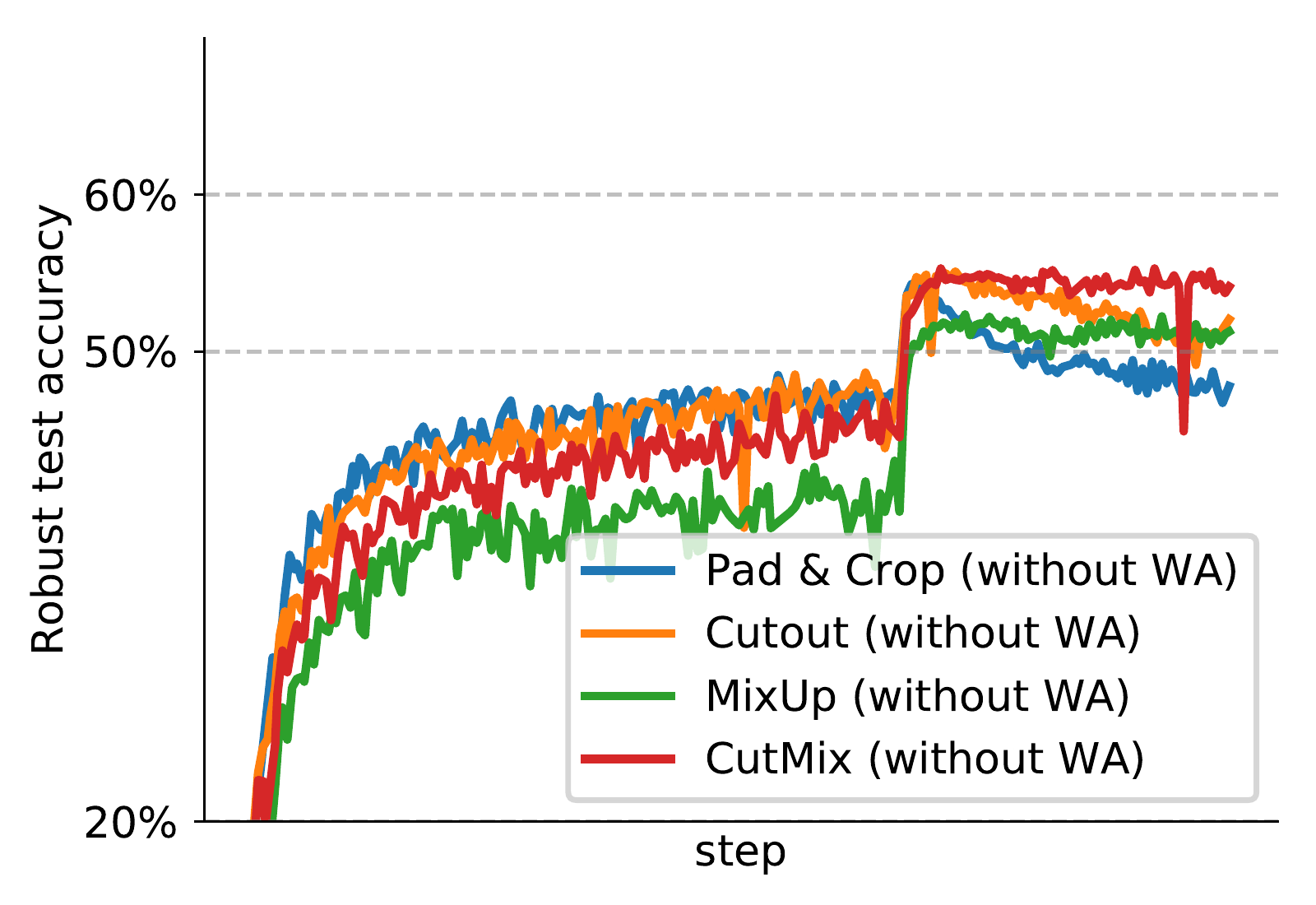}}
\hspace{1cm}
\subfigure[With WA]{\label{fig:augmentations_wa}\includegraphics[width=0.38\textwidth]{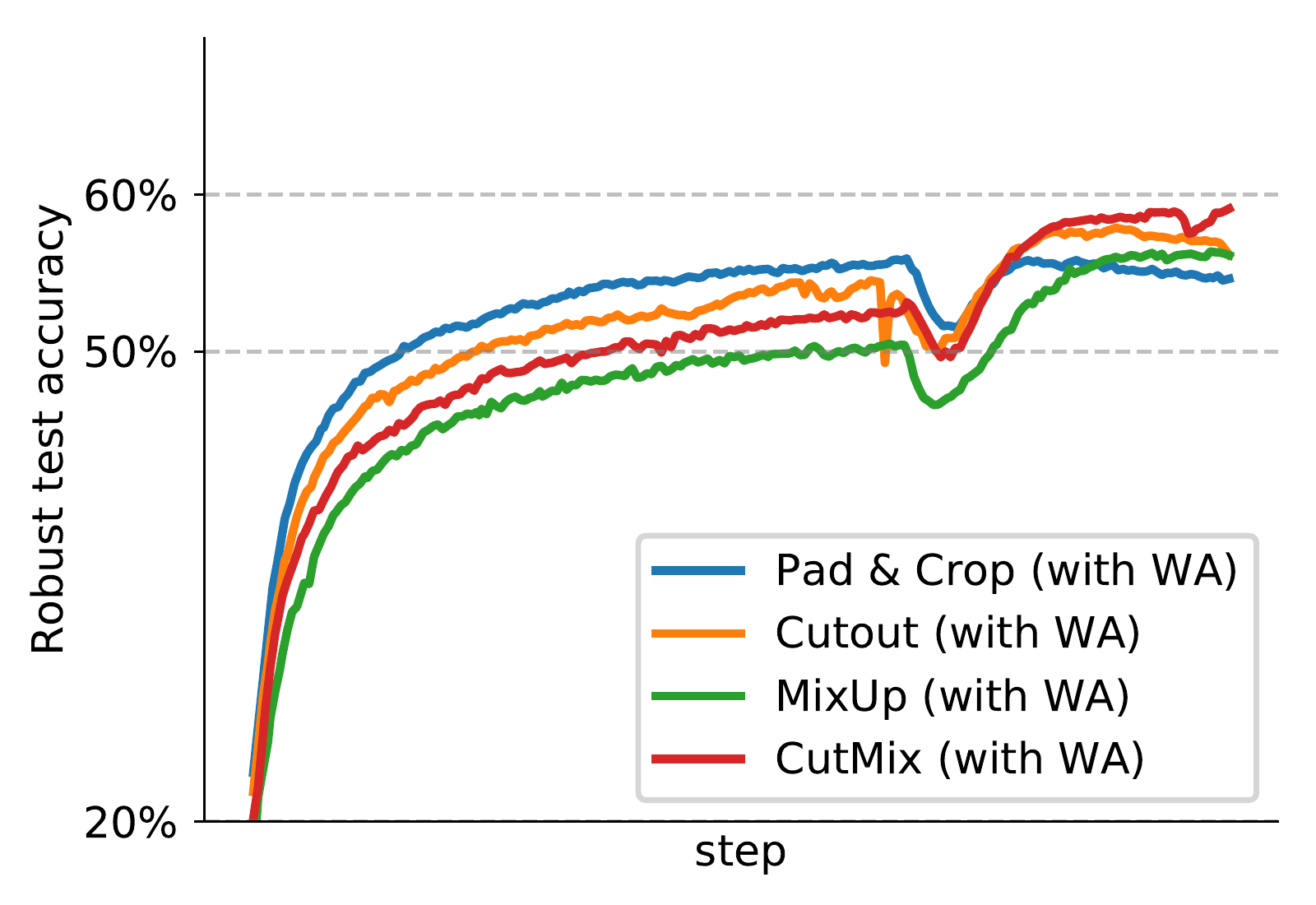}}
\caption{Accuracy against $\epsilon_\infty = 8/255$ on \cifar with and without using model weight averaging (WA) for different data augmentation schemes.
The model is a \wrn-28-10 and both panels show the evolution of the robust accuracy as training progresses (against \pgd{40}). The jump in robust accuracy two-thirds through training is due to a drop in learning rate.\label{fig:augmentations_wa_all}}
\end{figure*}

\citet{rice_overfitting_2020} show that combining data augmentation methods such as \emph{Cutout} or \emph{MixUp} with early stopping does not improve robustness upon early stopping alone.
While, these methods do not improve upon the ``best'' robust accuracy, they reduce the extent of robust overfitting, thus resulting in a slower decrease in robust accuracy compared to classical adversarial training (which uses random crops and weight decay).
This can be seen in \autoref{fig:augmentations_no_wa} where \emph{MixUp} without WA exhibits no decrease in robust accuracy, whereas the robust accuracy of the standard combination of \emph{random padding-and-cropping} without WA (\emph{Pad \& Crop}) decreases immediately after the change of learning rate.

\paragraph{Testing the hypothesis.}

Since \emph{MixUp} preserves robust accuracy while \emph{Pad \& Crop} does not, this comparison can be used to evaluate the hypothesis that WA is more beneficial when the performance between model iterations is maintained.
Therefore, we compare in \autoref{fig:augmentations_wa} the effect of WA on robustness when using \emph{MixUp}.
We observe that, when using WA, the performance of \emph{MixUp} surpasses the performance of \emph{Pad \& Crop}.
Indeed, the robust accuracy obtained by the averaged weights of \emph{Pad \& Crop} (in blue) slowly decreases after the change of learning rate, while the one obtained by \emph{MixUp} (in green) increases throughout training\footnote{The accuracy drop just after the change of learning rate stems from averaging very different weights.}.
Ultimately, \emph{MixUp} with WA obtains a higher robust accuracy despite the fact that the non-averaged \emph{MixUp} model has a significantly lower ``best'' robust accuracy than the non-averaged \emph{Pad \& Crop} model.
This finding is notable as it demonstrates for the first time the benefits of data augmentation schemes for adversarial training (this contradicts the findings from three recent publications: \cite{rice_overfitting_2020,wu2020adversarial,gowal_uncovering_2020}).


\paragraph{Exploring data augmentations.}

After verifying our hypothesis for \emph{MixUp}, we investigate if other augmentations can help maintain robust accuracy and also be combined with WA to improve robustness.
We concentrate on the following image patching techniques: \emph{Cutout}~\citep{devries2017improved} which inserts empty image patches and \emph{CutMix}~\cite{yun2019cutmix} which replaces part of an image with another.
In \autoref{sec:exp}, we also evaluate \emph{RICAP}~\citep{takahashi2018ricap} and \emph{SmoothMix}~\citep{lee2020smoothmix}. We describe more thoroughly these augmentations in \autoref{app:aug} where we also study additional augmentations with \emph{AutoAugment}~\citep{cubuk_autoaugment:_2018} and \emph{RandAugment}~\citep{cubuk2019randaugment}. Similarly to the analysis done for \emph{MixUp}, we report in \autoref{fig:augmentations_wa_all} the robust accuracy obtained by \emph{Cutout} and \emph{CutMix} with and without WA throughout training.
First, we note that these two techniques achieve a higher ``best'' robust accuracy than \emph{MixUp}, as shown in \autoref{fig:augmentations_no_wa}.
The ``best'' robust accuracy obtained by \emph{Cutout} and \emph{CutMix} is roughly identical to the one obtained by \emph{Pad \& Crop}, which is consistent with the results from \citet{rice_overfitting_2020}.
Second, while \emph{Cutout} suffers from robust overfitting, \emph{CutMix} does not.
Hence, as demonstrated in the previous sections, we expect WA to be more useful with \emph{CutMix}.
Indeed, we observe in \autoref{fig:augmentations_wa} that the robust accuracy of the averaged model trained with \emph{CutMix} keeps increasing throughout training and that its maximum value is significantly above the best accuracy reached by the other augmentation methods.
In \autoref{sec:exp}, we conduct thorough evaluations of these methods against stronger attacks.


\section{Experimental setup} 
\label{sec:exp_setup}

\paragraph{Architecture.}

We use \glspl*{wrn}~\citep{he2015deep,zagoruyko2016wide} as our backbone network.
This is consistent with prior work \citep{madry_towards_2017,rice_overfitting_2020,zhang_theoretically_2019,uesato_are_2019,gowal_uncovering_2020} which use diverse variants of this network family.
Furthermore, we adopt the same architecture details as \citet{gowal_uncovering_2020} with Swish/SiLU~\citep{hendrycks2016gaussian} activation functions.
Most of the experiments are conducted on a \wrn-28-10 model which has a depth of 28, a width multiplier of 10 and contains 36M parameters.
To evaluate the effect of data augmentations on wider and deeper networks, we also run several experiments using \wrn-70-16, which contains 267M parameters. 

\paragraph{Outer minimization.}
We use TRADES~\citep{zhang_theoretically_2019} optimized using SGD with Nesterov momentum~\citep{polyak1964some, nesterov27method} and a global weight decay of $5 \times 10^{-4}$. 
We train for $400$ epochs with a batch size of $512$ split over $32$ Google Cloud TPU v$3$ cores~\cite{bradbury_jax_2018}, and the learning rate is initially set to 0.1 and decayed by a factor 10 two-thirds-of-the-way through training.
We scale the learning rates using the linear scaling
rule of \citet{goyal2017accurate} (i.e., $\textrm{effective LR} = \max(\textrm{LR} \times \textrm{batch size} / 256, \textrm{LR})$).
The decay rate of WA is set to $\tau=0.999$.
With these settings, training a \wrn-28-10 takes on average 2.5 hours.

\paragraph{Inner minimization.}
Adversarial examples are obtained by maximizing the  Kullback-Leibler  divergence between the predictions made on clean inputs and those made on adversarial inputs~\citep{zhang_theoretically_2019}.
This optimization procedure is done using the Adam optimizer \citep{kingma_adam:_2014} for 10 \gls*{pgd} steps.
We take an initial step-size of $0.1$ which is then decreased to $0.01$ after 5 steps.

\paragraph{Evaluation.}

We follow the evaluation protocol designed by \citet{gowal_uncovering_2020}.
Specifically, we train two (and only two) models for each hyperparameter setting, perform early stopping for each model on a separate validation set of 1024 samples using \pgd{40} similarly to~\citet{rice_overfitting_2020} and pick the best model by evaluating the robust accuracy on the same validation set .
Finally, we report the robust test accuracy against a mixture of \autoattack~\citep{croce_reliable_2020} and \multitargeted~\citep{gowal_alternative_2019}, which is denoted by \textsc{AA+MT}.
This mixture consists in completing the following sequence of attacks: \autopgd on the cross-entropy loss with 5 restarts and 100 steps, \autopgd on the difference of logits ratio loss with 5 restarts and 100 steps and finally \multitargeted on the margin loss with 10 restarts and 200 steps.
The training curves, such as those visible in \autoref{fig:robust_overfitting}, are always computed using \gls{pgd} with 40 steps and the Adam optimizer (with step-size decayed by 10$\times$ at step 20 and 30).

\section{Experimental results} 
\label{sec:exp}

First, we will investigate which augmentation techniques benefit the most from WA and why. Then, we will generalize our approach to other architecture, threat model and datasets. Finally, we provide empirical evidence showing that WA exploits data augmentation by ensembling model snapshots which differ at the individual prediction level.

\begin{figure}
\begin{floatrow}
\ffigbox{%
  \includegraphics[height=.6\linewidth]{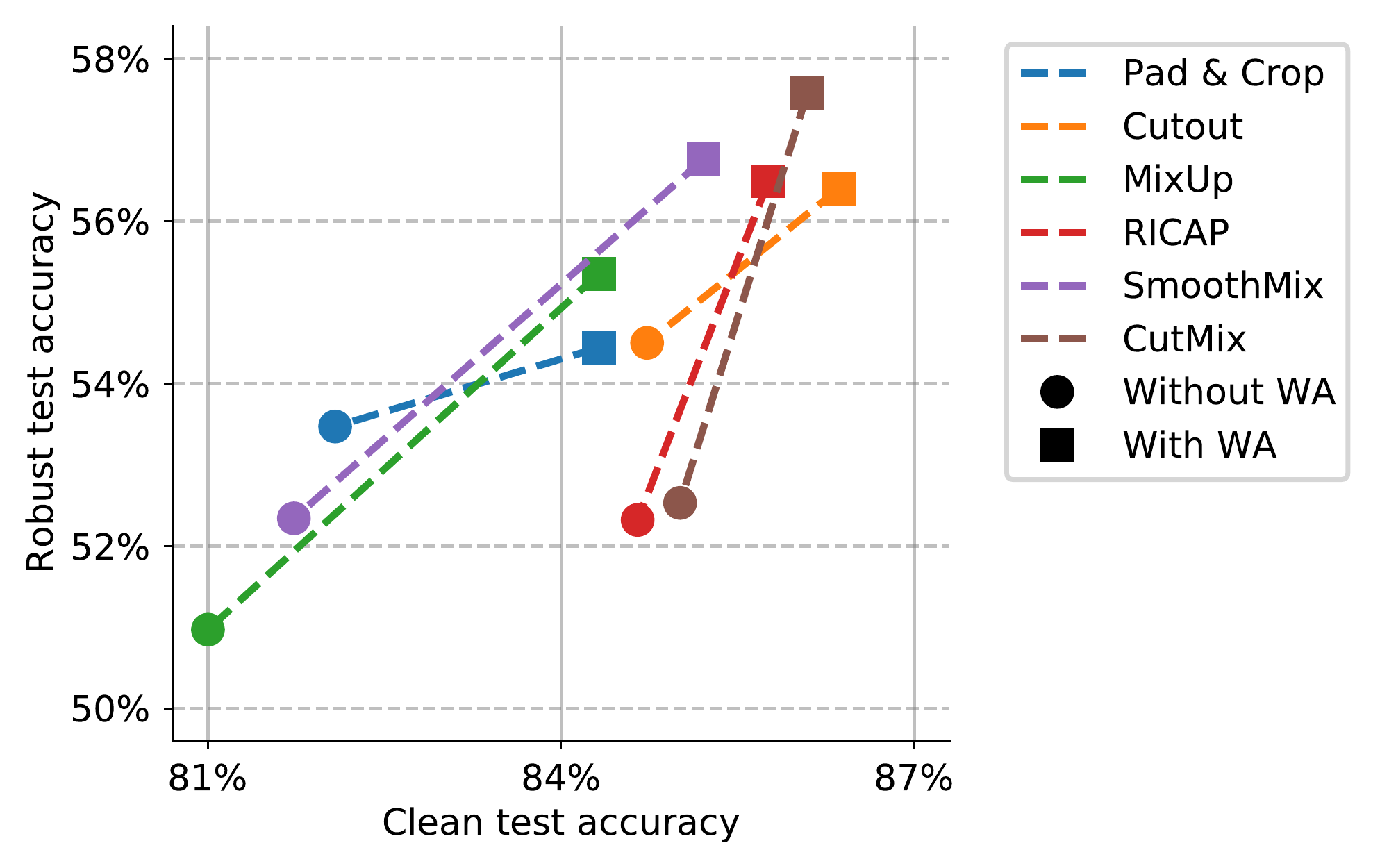}
}{%
  \caption{Clean (without adversarial attacks) accuracy and robust accuracy (against \textsc{AA+MT}) for a \wrn-28-10 trained against $\epsilon_\infty = 8/255$ on \cifar for different data augmentation techniques. The lines from circles to squares represent the performance change obtained when using WA. \label{fig:augmentations_summary}}%
}
\ffigbox{%
  \includegraphics[height=.6\linewidth]{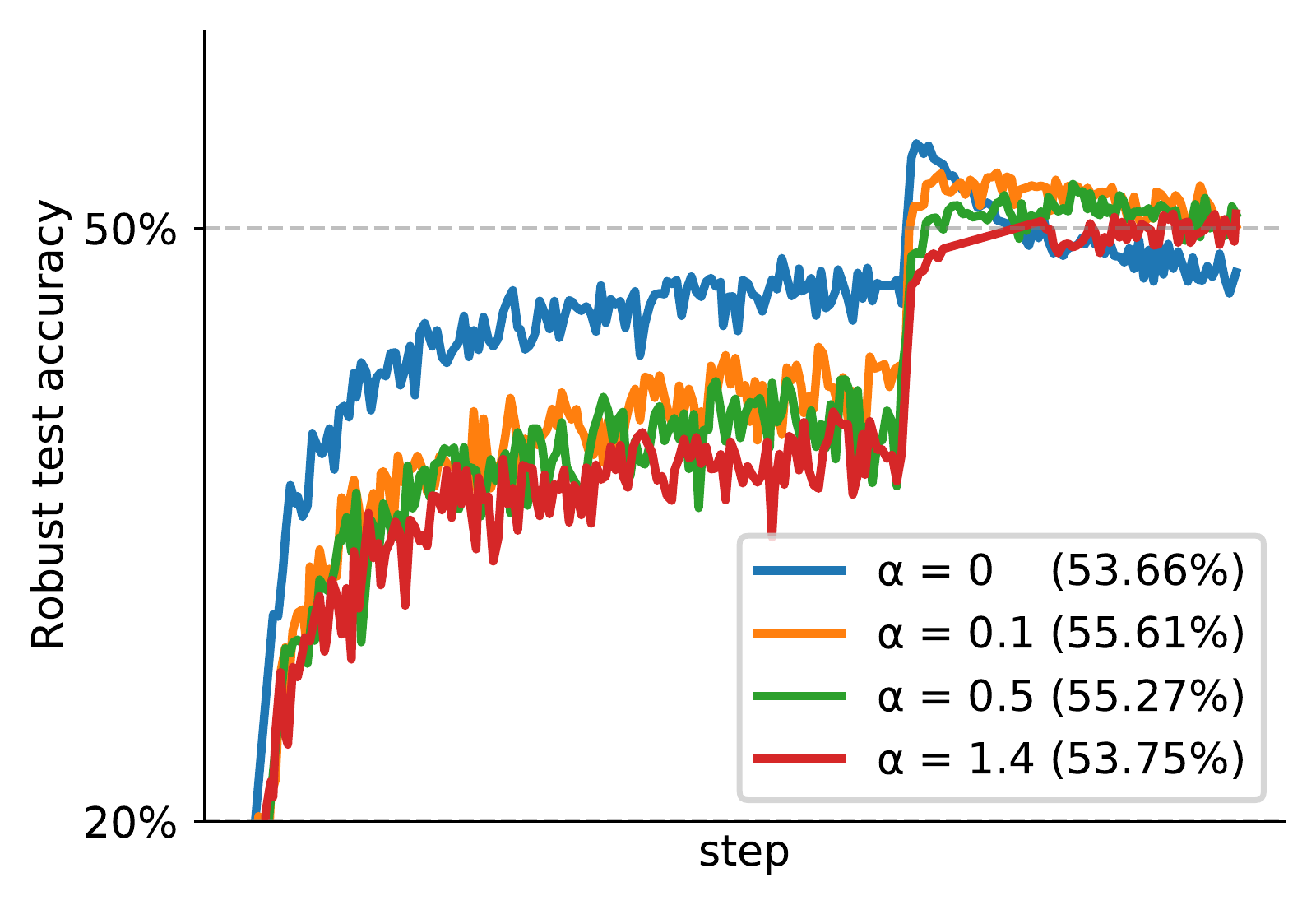}
}{%
  \caption{The graph shows the robust test accuracy against \pgd{40} with $\epsilon_\infty = 8/255$ on \cifar without using WA as we vary the mixing rate $\alpha$ of \emph{MixUp}. We report in the legend the robust accuracy (against \textsc{AA+MT}) after applying weight averaging to the corresponding runs.\label{fig:mixup_alpha_pgd}}%
}
\end{floatrow}
\end{figure}

\subsection{Comparing data augmentations}

Here, we compare data augmentations with and without WA. We consider as baseline the \emph{Pad \& Crop} augmentation which reproduces the current state-of-the-art set by \citet{gowal_uncovering_2020}. This augmentation consists in first padding the image by 4 pixels on each side and then taking a random $32\times 32$ crop.
In \autoref{fig:augmentations_summary}, we compare this baseline with various data augmentations, \emph{MixUp}, \emph{Cutout}, \emph{CutMix}, \emph{RICAP} and \emph{SmoothMix}.
A first cluster (the four top squares), containing \emph{RICAP}, \emph{Cutout}, \emph{SmoothMix} and \emph{CutMix}, includes the four methods that occlude local information with patching and provide a significant boost upon the baseline with +3.06\% in robust accuracy for \emph{CutMix} and an average improvement of +1.54\% in clean accuracy.
The other cluster, with \emph{MixUp}, only improves the robust accuracy upon the baseline by a small margin of +0.91\%. Furthermore, we also point out that \emph{Pad \& Crop} and \emph{Cutout}, which were the two augmentations suffering from robust overfitting in \autoref{fig:augmentations_no_wa}, benefit the least of WA in \autoref{fig:augmentations_summary} (smaller vertical gains). This is consistent with our hypothesis of \autoref{sec:hypothesis} that WA is the most beneficial when robust overfitting is reduced.

\begin{figure*}[t]
\centering
\subfigure[\label{fig:mixup_alpha}]{\includegraphics[width=.32\columnwidth]{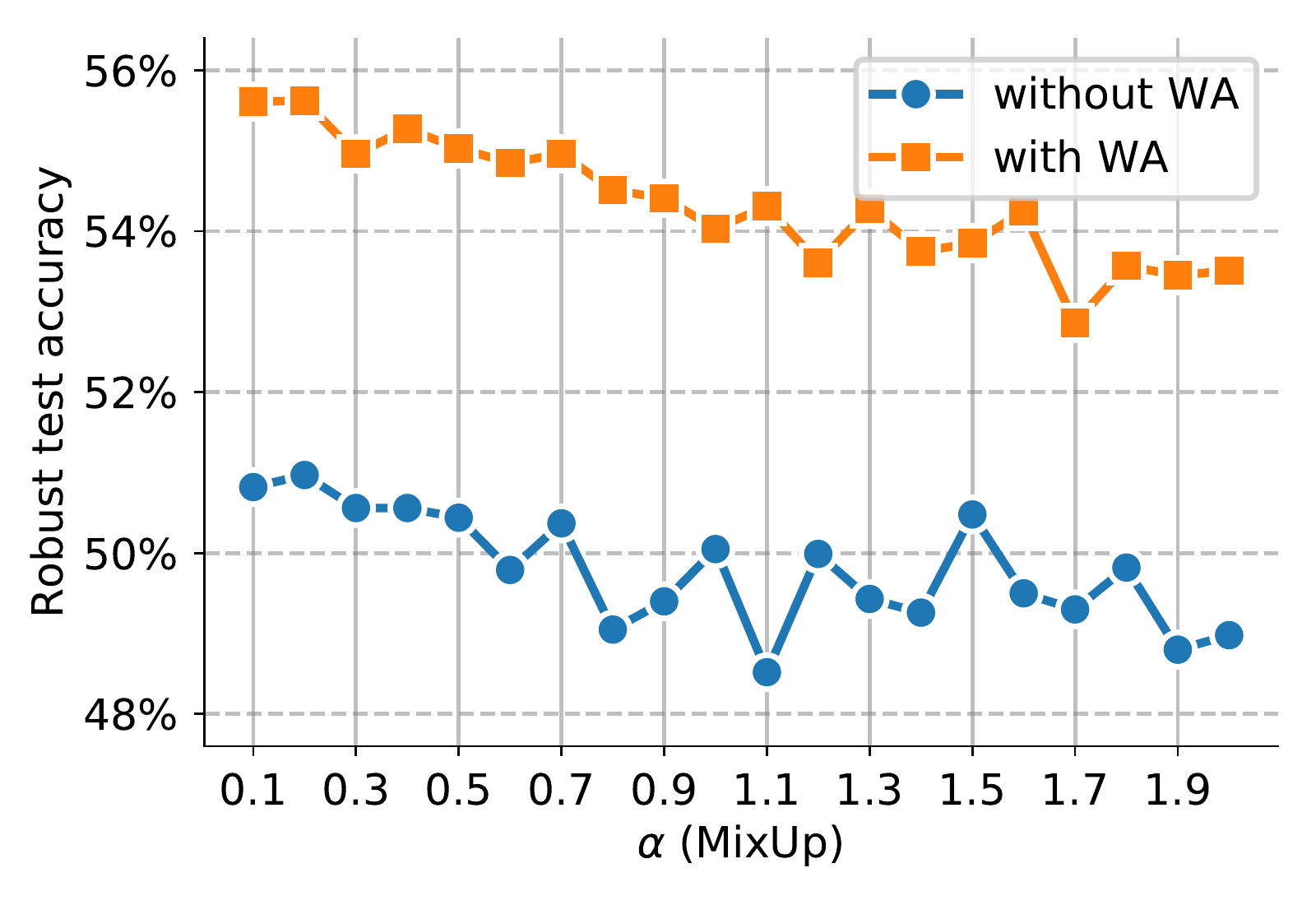}}
\subfigure[\label{fig:cutout_length}]{\includegraphics[width=.32\columnwidth]{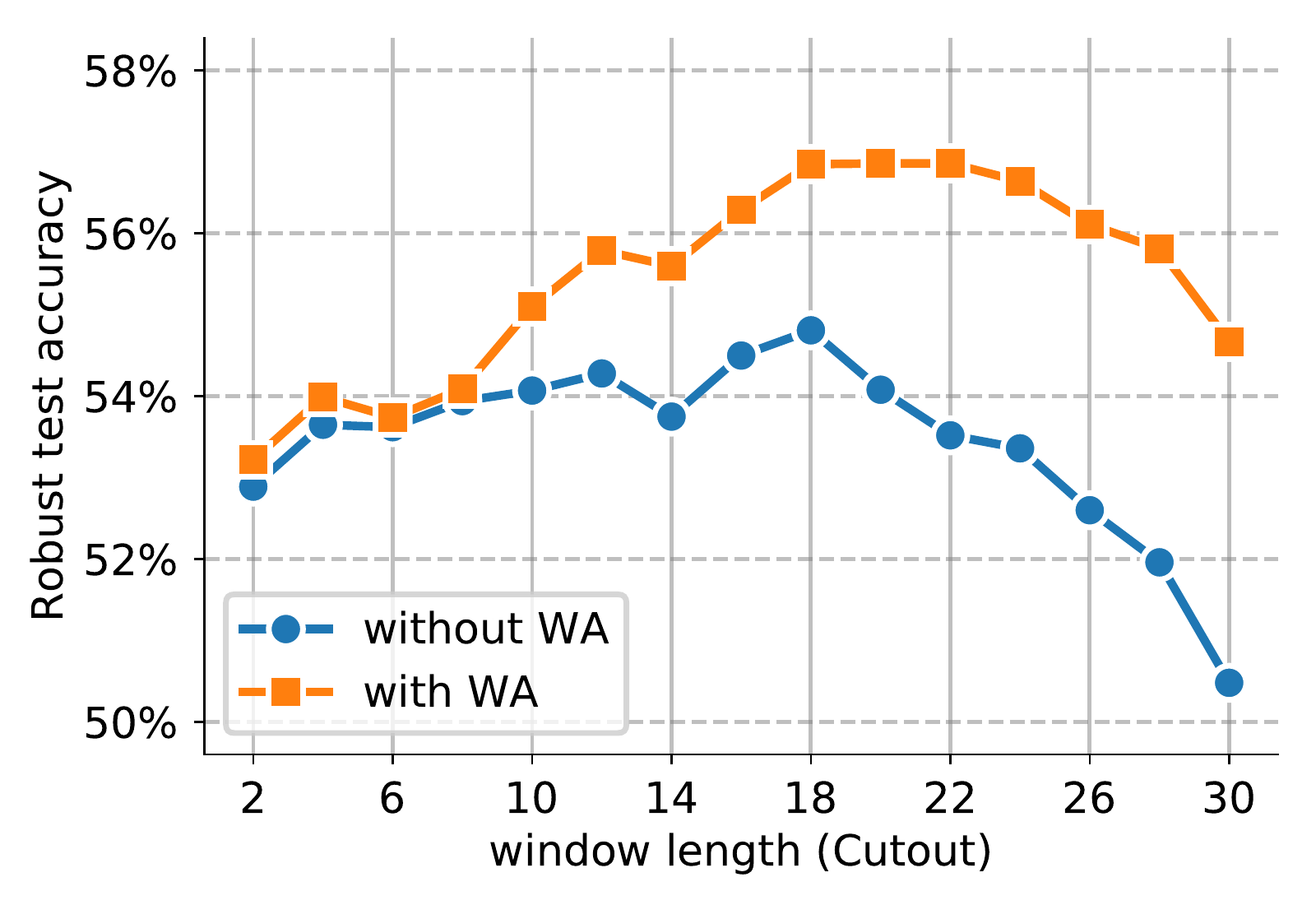}}
\subfigure[\label{fig:cutmix_length}]{\includegraphics[width=.32\columnwidth]{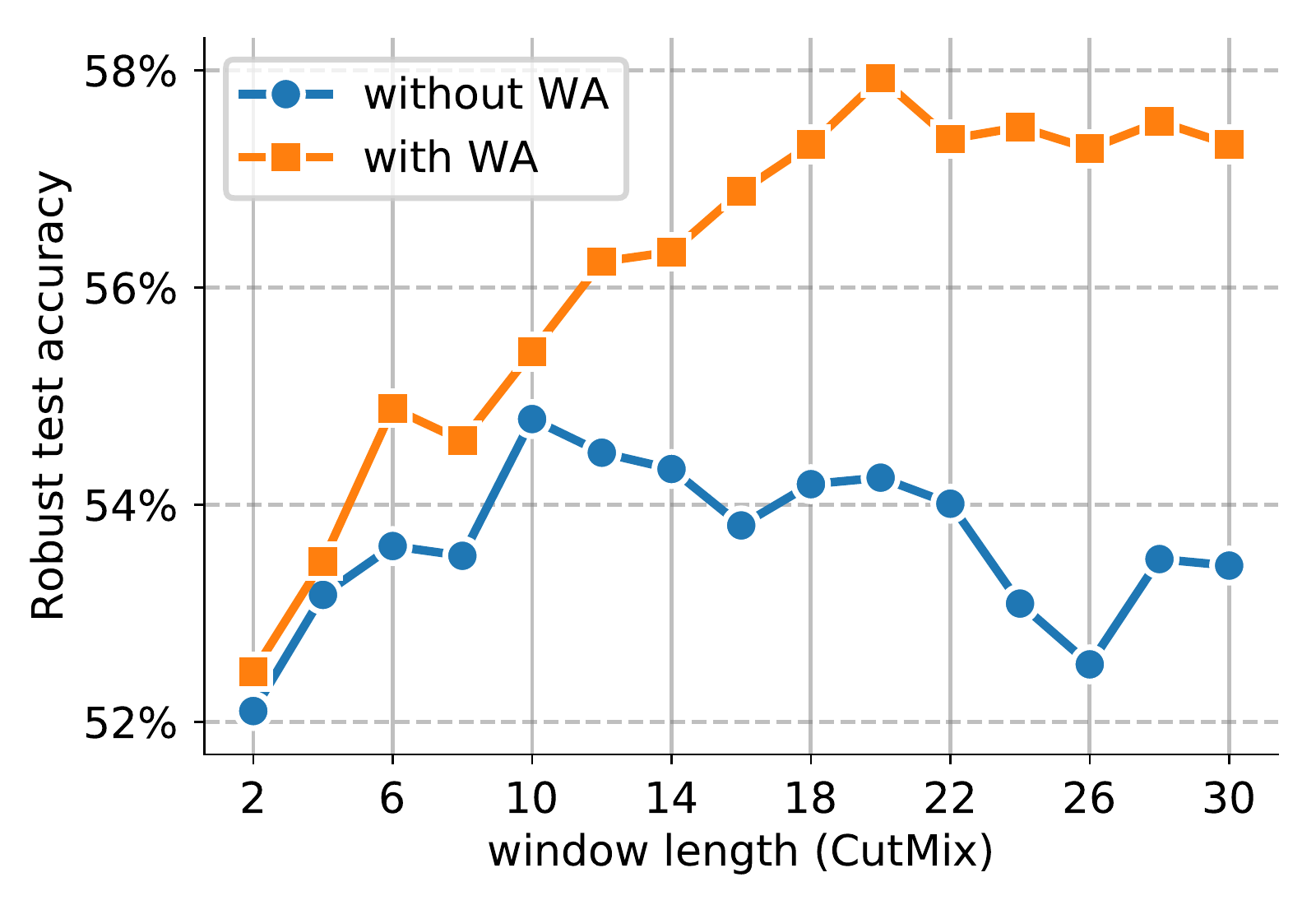}}
\caption{Robust test accuracy against AA+MT with $\epsilon_\infty= 8/255$ on \cifar as we vary \subref{fig:mixup_alpha} the mixing rate $\alpha$ of \emph{MixUp}, \subref{fig:cutout_length} the window length when using \emph{Cutout} and \subref{fig:cutmix_length} the window length when using \emph{CutMix}.  The model is a \wrn-28-10 and we compare the settings without and with WA. As a reference when training only with \emph{Pad \& Crop}, the same model with WA and without WA reaches 54.44\% and 53.66\% robust accuracy, respectively. Similarly, without any augmentation, the models with WA and without WA achieve 49.74\% and 42.27\%, respectively.}
\label{fig:ablation_sweeps}
\end{figure*}

\paragraph{\emph{MixUp}.}
A possible explanation to the worse performance of \emph{MixUp} lies in the fact that \emph{MixUp}, which samples the image mixing weight with a beta distribution $\operatorname{Beta}(\alpha, \alpha)$, tends to either produce images that are far from the original data distribution (when $\alpha$ is large) or too close to the original samples (when $\alpha$ is small).
In fact, \autoref{fig:mixup_alpha_pgd}, which shows the robust accuracy when training without WA, illustrates the trade-off between robust overfitting and underfitting as increasing $\alpha$ can lead to robust underfitting (red curve) while an $\alpha$ too close to 0 would lead to robust overfitting.
More specifically, we show in \autoref{fig:mixup_alpha} that \emph{MixUp}'s robust accuracy against AA+MT keeps decreasing as $\alpha$ increases and the best performance with WA is reached at $\alpha=0.2$, corresponding to blended images close to the original images.

\paragraph{Spatial composition techniques.}
\autoref{fig:ablation_sweeps}(b,c) show the robust test accuracy as we vary the window length of the patches when using \emph{Cutout} and \emph{CutMix}\footnote{For all CutMix results except those reported in Fig 6(c)), the window length is randomly sampled such that the patch area ratio follows a beta distribution with parameters $\alpha = 1$ and $\beta = 1$.}. We observe that these two techniques are the most beneficial when using large window lengths with a peak reached at a length of 20. Hence, contrary to \emph{MixUp}, they work best with patched images which greatly differ from the original images.
This performance gap between \emph{MixUp} and \emph{Cutout}/\emph{CutMix} illustrates a notable difference in the use of data augmentation for adversarial training compared to nominal training. Indeed, adversarial training can lead to underfitting.
This leads to some augmentation techniques working better than others in the context of adversarial training.
This is the case for spatial composition techniques which outperform blending techniques like \emph{MixUp}.
A possible explanation is that low-level features tend to be destroyed by \emph{MixUp}, whereas composition techniques locally maintain these low-level features. Hence, we hypothesize that augmentations designed for robustness need to preserve low-level features. We provide further evidence to support this hypothesis in~\autoref{app:aug} by showing that some components of \emph{RandAugment} such as \emph{Posterize} or \emph{Invert} are detrimental to adversarial robustness.

\subsection{Generalizing to other architectures, threat model and datasets}

\paragraph{Generalizing to other architectures.} Table \ref{table:pad_vs_cutmix} shows the performance of \emph{CutMix} and the \emph{Pad \& Crop} baseline when varying the model architecture and size. 
We experiment with different variants of WideResNet and ResNet.
We use WA for both \emph{CutMix} and the \emph{Pad \& Crop} baseline.
We observe that \emph{CutMix} consistently outperforms the baseline by at least +2.90\% in robust accuracy across all model sizes for WideResNet and by at least +1.76\% for ResNet models.

\paragraph{Generalizing to another threat model.} 
We extend our evaluation to \ltwo-norm bounded perturbations.
Table \ref{tab:combining_gen_aug} shows the performance of data augmentation on \cifar against $\epsilon_\infty = 8/255$ and $\epsilon_2 = 128/255$.
We observe that using \emph{CutMix}  provides a significant boost in robust accuracy for both threat models with up to +2.93\% (in the \linf setting) and +2.16\% (in the \ltwo setting).

\paragraph{Generalizing to other datasets.} To evaluate the generality of our approach, we evaluate it on \cifarh~\cite{krizhevsky2009learning}, \svhn~\cite{netzer2011reading} and \tinyimagenet~\cite{russakovsky2015imagenet} and we report the results in Table \ref{table:datasets_results}.
First, on \cifarh, our best model reaches 32.43\% against \autoattack and improves noticeably on the state-of-the-art by +2.40\% (in the setting that does not use any external data).
Second, on \tinyimagenet with a \wrn-28-10 we obtain a significant +2.00\% boost for robust accuracy against \textsc{AA+MT} with $\epsilon_\infty = 8/255$.
Finally, on \svhn, our best model reaches 57.32\% against \textsc{AA+MT} and improves on the baseline by a smaller margin than on \cifar, \cifarh or \tinyimagenet. This smaller improvement is expected as \emph{CutMix} is not suited to \svhn because images of \svhn contain multiple digits per image.

\begin{figure}[t]
\begin{floatrow}
\capbtabbox{%
\resizebox{0.41\textwidth}{!}{%
  \begin{tabular}{l|cc|cc}
    \hline
    \cellcolor{header} & \multicolumn{2}{c|}{\cellcolor{header} \textsc{Pad \& Crop}} & \multicolumn{2}{c}{\cellcolor{header} \textsc{CutMix}} \Tstrut \\
    \cellcolor{header} \textsc{Setup} & \cellcolor{header} \textsc{Clean} & \cellcolor{header} \textsc{Robust} & \cellcolor{header} \textsc{Clean} & \cellcolor{header} \textsc{Robust} \Bstrut \\
    \hline
    \hline
    \multicolumn{5}{l}{\cellcolor{subheader} \textsc{Varying the architecture}} \TBstrut \\
    \hline
    ResNet-18 & 83.12\% & 50.52\% & 80.57\%  & \textbf{52.28\%} \Tstrut \\
    ResNet-34 & 84.68\% & 52.52\% & 83.35\%  & \textbf{54.80\%} \Bstrut \\
    \hline
    \hline
    \wrn-28-10 & 84.32\% & 54.44\% & 86.09\%  & \textbf{57.50\%} \Tstrut \\
    \wrn-34-10 & 84.89\% & 55.13\% & 86.18\%  & \textbf{58.09\%} \rule{0pt}{0.0ex} \\
    \wrn-34-20 & 85.80\% & 55.69\% & 87.80\%  & \textbf{59.25\%} \rule{0pt}{0.0ex} \\
    \wrn-70-16 & 86.02\% & 57.17\% & 87.25\%  & \textbf{60.07\%} \Bstrut \\
    \hline
\end{tabular}%
}
}{%
  \caption{Robust test accuracy (against \textsc{AA+MT}) against $\epsilon_\infty = 8/255$ on \cifar for different architectures. In all cases, we use weight averaging and we compare \emph{Pad \& Crop} and \emph{CutMix}.\label{table:pad_vs_cutmix}}%
}
\capbtabbox{%
\resizebox{0.53\textwidth}{!}{%
\begin{tabular}{l|cc|cc}
    \hline
     \cellcolor{header}   & \multicolumn{2}{c|}{\cellcolor{header} \linf} & \multicolumn{2}{c}{\cellcolor{header} \ltwo}  \Tstrut \\
     \cellcolor{header} \textsc{Setup} & \cellcolor{header} \textsc{Clean} & \cellcolor{header} \textsc{Robust} & \cellcolor{header} \textsc{Clean} & \cellcolor{header} \textsc{Robust} \Bstrut \\
    \hline
    \hline
    \multicolumn{5}{l}{\cellcolor{subheader} \textsc{\wrn-28-10} } \TBstrut \\
    \hline
    \citet{gowal_uncovering_2020} (trained by us)  & 84.32\% & 54.44\% & 88.60\% & 72.56\% \Tstrut \\
    Ours (CutMix) & 86.22\% & \textbf{57.50\%} & 91.35\% & \textbf{76.12\%}  \Bstrut\\
    \hline
    \hline
    \multicolumn{5}{l}{\cellcolor{subheader} \textsc{\wrn-70-16}} \TBstrut \\
    \hline
    \citet{gowal_uncovering_2020} (trained by us)  & 85.29\% & 57.14\% & 90.90\% & 74.50\%   \Tstrut \\
    Ours (CutMix) & 87.25\% & \textbf{60.07\%} & 92.43\% & \textbf{76.66\%}  \Bstrut\\
    \hline
    \end{tabular}
}
}{%
  \caption{Clean (without adversarial attacks) accuracy and robust accuracy (against \textsc{AA+MT}) on \cifar as we both test against $\epsilon_\infty = 8/255$ and $\epsilon_2 = 128/255$.\label{tab:combining_gen_aug}}%
}
\end{floatrow}
\end{figure}

\begin{figure}
\floatbox[{\capbeside\thisfloatsetup{capbesideposition={right,top},capbesidewidth=6cm}}]{table}[\FBwidth]
{
\caption{Clean and robust accuracy (\textsc{AA+MT} and \autoattack for select models) on \cifarh, \svhn and \tinyimagenet against $\epsilon_\infty = 8/255$ obtained by different models (with WA). The 'retrained' indication means that the models have been retrained according to \citet{gowal_uncovering_2020}'s methodology.\label{table:datasets_results}}
}
{
\resizebox{.46\textwidth}{!}{
\begin{tabular}{l|ccc}
    \hline
    \cellcolor{header} \textsc{Model} & \cellcolor{header} \textsc{Clean} & \cellcolor{header} \textsc{AA+MT} & \cellcolor{header} \textsc{AA} \TBstrut \\
    \hline
    \hline
    \multicolumn{4}{l}{\cellcolor{subheader} \textsc{\cifarh} } \TBstrut \\
    \hline
    \citet{cui2020learnable} (\wrn-34-10) & 60.64\% & -- & 29.33\% \Tstrut \\
    \wrn-28-10 (retrained) & 59.05\% & 28.75\% & -- \\
    \wrn-28-10 (CutMix) & 62.97\% & \textbf{30.50\%} & \textbf{29.80\%} \Bstrut\\
    \hline
    \citet{gowal_uncovering_2020} (\wrn-70-16) & 60.86\% & 30.67\% & 30.03\% \Tstrut \\
    \wrn-70-16 (retrained) & 59.65\% & 30.62\% & -- \\
    \wrn-70-16 (CutMix) & 65.76\% & \textbf{33.24\%} & \textbf{32.43\%} \Bstrut\\
    \hline
    \hline
    \multicolumn{4}{l}{\cellcolor{subheader} \textsc{\svhn} } \TBstrut \\
    \hline
    \wrn-28-10 (retrained) & 92.87\% & 56.83\% & -- \Tstrut \\
    \wrn-28-10 (CutMix) & 94.52\% & \textbf{57.32\%} & -- \Bstrut\\
    \hline
    \hline
    \multicolumn{4}{l}{\cellcolor{subheader} \textsc{\tinyimagenet} } \TBstrut \\
    \hline
    \wrn-28-10 (retrained) & 53.27\% & 21.83\% & -- \Tstrut \\
    \wrn-28-10 (CutMix) & 53.69\% & \textbf{23.83\%} & -- \Bstrut\\
    \hline
\end{tabular}
}
}
\end{figure}

\subsection{Empirical elements on how weight averaging exploits data augmentation}

\paragraph{Motivating model ensembling.}
First, we show that model ensembling can be used to improve robust accuracy. To do so, we evaluate ensembling early-stopped models which have been trained from scratch independently. We ensemble two early-stopped \wrn-28-10 models trained on \cifar with \emph{Pad \& Crop} by taking the average of the two independent models at the prediction level. In spite of this naive ensembling approach, we observe a significant boost in robust performance as this ensemble reaches 55.69\% robust accuracy against \textsc{AA+MT} compared to 54.44\% with a single model. Hence, even a simple ensemble of two independent runs can exploit the variance in individual robust predictions. Actually, the boost in robust accuracy is even stronger when ensembling two early-stopped \wrn-28-10 models trained with \emph{CutMix} as the ensemble reaches 56.35\% robust accuracy which is +3.82\% better compared to an individual model. Augmentation techniques such as \emph{CutMix} promote more diversity between runs than \emph{Pad \& Crop}, leading thereby to better robust performance when ensembling. This is further evidence that ensembling by its ability of exploiting the diversity of the models is mainly responsible for robustness improvements.

\begin{figure}
\begin{floatrow}
\ffigbox{%
  \includegraphics[height=.4\linewidth]{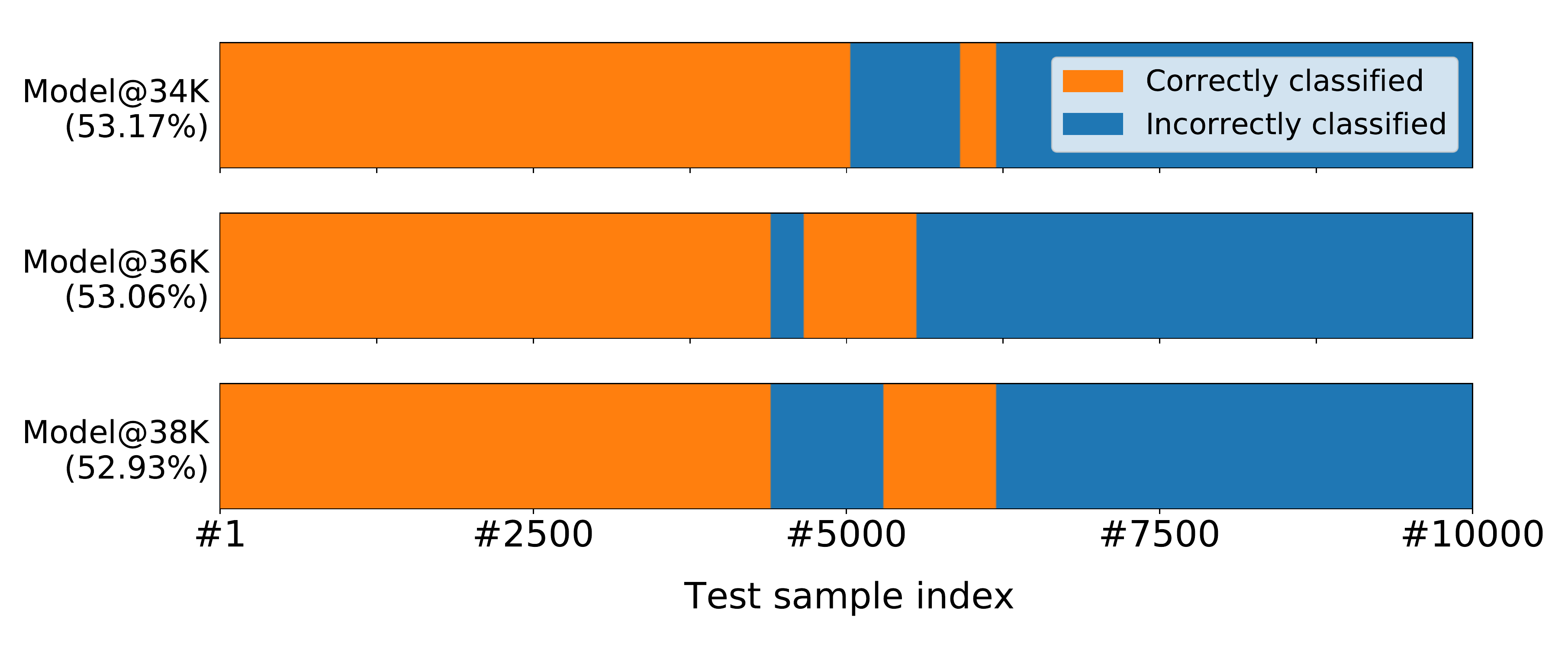}
}{%
   \caption{
The bar plots show the outcome of each individual robust prediction for different snapshots of a same training run of a \wrn-28-10 against $\epsilon_\infty = 8/255$ on \cifar without model weight averaging.
The test sample indices have been re-ordered such as to show contiguous blocks. The plots show a significant variation in individual robust predictions across different snapshots while the total robust accuracy (i.e. the number in parenthesis) remains stable.
\label{fig:chromo}
}
}
\ffigbox{%
  \includegraphics[height=.5\linewidth]{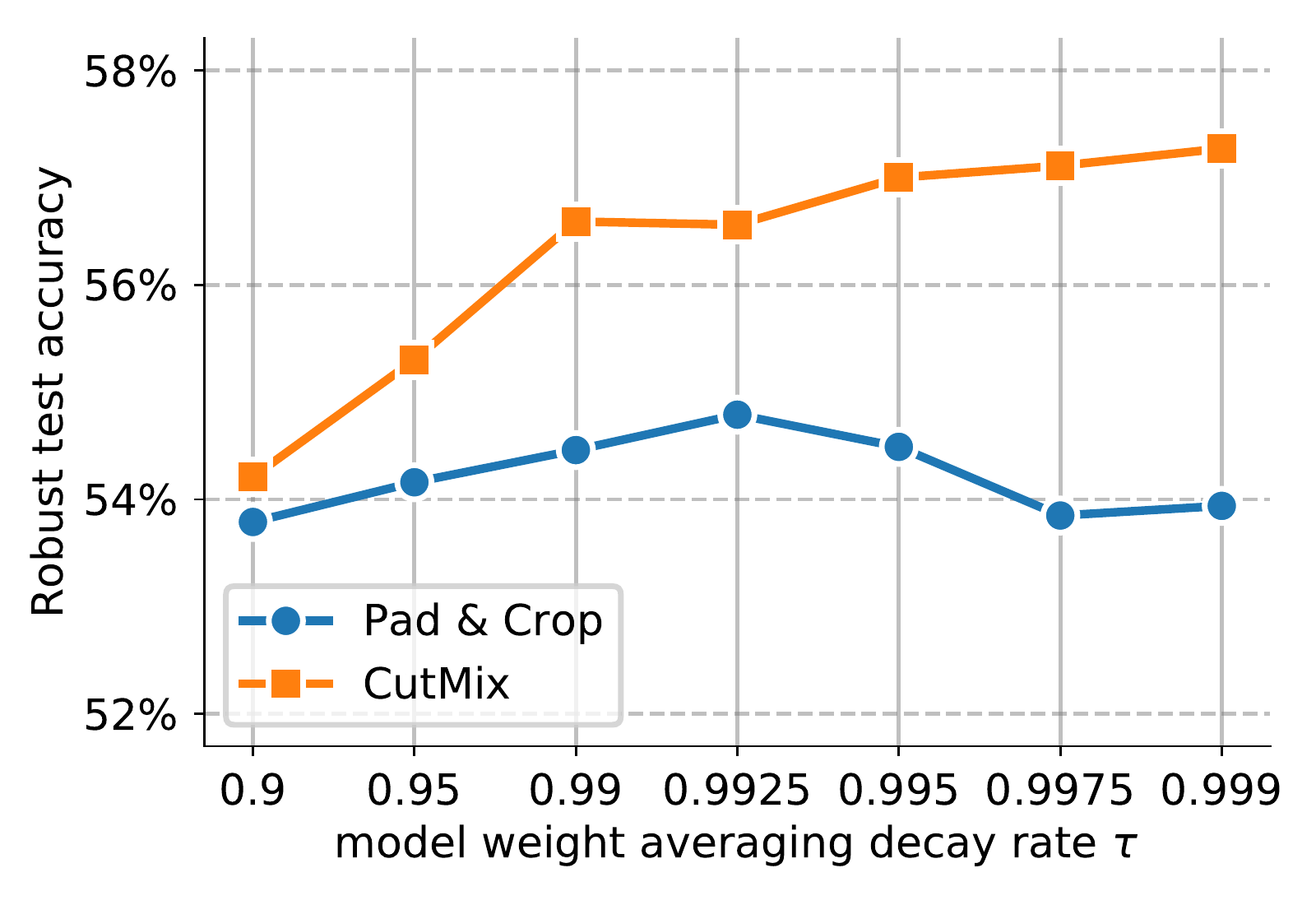}
}{%
    \caption{Robust test accuracy against \textsc{AA+MT} with $\epsilon_\infty= 8/255$ on \cifar as we vary the decay rate of the model weight averaging. The model is a \wrn-28-10, which is trained either with \emph{CutMix} or \emph{Pad \& Crop}.\label{fig:WA_decay}}%
}
\end{floatrow}
\end{figure}

\paragraph{Model ensembling by weight averaging.}
We would like to ensemble more than two models but it would be inefficient computationally and memory wise to average the predictions of many independently trained models. 
To circumvent this issue, the naive ensembling approach is replaced by model weight averaging as we exploit the commonly known fact~\citep{gowal_alternative_2019,qin_adversarial_2019,chen2021robust} that models trained with adversarial training tend to be locally linear.
Indeed, under the assumption of linearity, weight averaging becomes equivalent to model ensembling. 
Hence, instead of ensembling independently trained models, weight averaging ensembles model iterations obtained during one training run. 
As discussed in the previous paragraph, model ensembling improves robustness by exploiting the diversity of equally performing models so we need the model iterations used with weight averaging to have similar robust performance but also some diversity in individual robust predictions. 
As we have previously seen in \autoref{fig:augmentations_no_wa}, \emph{CutMix} without weight averaging prevents robust overfitting and leads to a flat robust accuracy after the change of learning rate. Hence, these model snapshots share the same total accuracy but we would like to know if they differ at the individual prediction level.
\autoref{fig:chromo} represents the individual robust predictions on test samples for three snapshots taken during training. While these three snapshots roughly have the same number of correctly classified samples, we see that there are on average 888 errors (out of 10k samples) per snapshot which are not made in the other two snapshots. 
This shows that augmentations which avoid robust overfitting such as \emph{CutMix} produce diverse and equally performing model iterations, which can be ensembled effectively by model weight averaging, leading thereby to improved robust performance.

\paragraph{The limits when exploiting the diversity between model iterations.}
When robust overfitting occurs, weight averaging is still helpful but to a lesser extent. In fact, a compromise must be found between the performance boost from ensembling diverse model iterations and the performance loss from incorporating model iterations with degraded performance in the ensemble.
We illustrate this point by running an ablation study in \autoref{fig:WA_decay} measuring the robust accuracy obtained when varying the decay rate $\tau$ of model weight averaging and using either \emph{Pad \& Crop} or \emph{CutMix}.
While for \emph{CutMix} increasing the weight averaging decay rate (i.e. ensembling more model iterations) always results in better robust performance, we observe that for \emph{Pad \& Crop} the maximum robust performance is obtained at $\tau = 0.9925$. When the weight averaging decay rate becomes too large ($\tau>0.9925$), too many model iterations with degraded performance are incorporated in the ensemble, thus hurting the robust performance of the ensemble. Hence, the diversity between model iterations can only compensate up to a certain point for the decrease in robust performance due to robust overfitting.


\section{Conclusion}

Contrary to previous works~\citep{rice_overfitting_2020,gowal_uncovering_2020,wu2020adversarial}, which have tried data augmentation techniques to train adversarially robust models without success, we demonstrate that combining data augmentations with model weight averaging can significantly improve robustness.
We also provide insights on why weight averaging works better with data augmentations which reduce robust overfitting. We show in fact that model snapshots of a same run have the same total robust accuracy but they greatly differ at the individual prediction level, thus allowing a performance boost when ensembling these snapshots. Code and models are available online at \url{https://github.com/deepmind/deepmind-research/tree/master/adversarial_robustness}.

\bibliography{references}

\begin{thebibliography}{62}
\providecommand{\natexlab}[1]{#1}
\providecommand{\url}[1]{\texttt{#1}}
\expandafter\ifx\csname urlstyle\endcsname\relax
  \providecommand{\doi}[1]{doi: #1}\else
  \providecommand{\doi}{doi: \begingroup \urlstyle{rm}\Url}\fi

\bibitem[Andriushchenko et~al.(2020)Andriushchenko, Croce, Flammarion, and
  Hein]{andriushchenko_square_2019}
M.~Andriushchenko, F.~Croce, N.~Flammarion, and M.~Hein.
\newblock Square {Attack}: a query-efficient black-box adversarial attack via
  random search.
\newblock \emph{Eur. Conf. Comput. Vis.}, 2020.

\bibitem[Athalye and Sutskever(2018)]{athalye_synthesizing_2017}
A.~Athalye and I.~Sutskever.
\newblock Synthesizing robust adversarial examples.
\newblock \emph{Int. Conf. Mach. Learn.}, 2018.

\bibitem[Athalye et~al.(2018)Athalye, Carlini, and
  Wagner]{athalye_obfuscated_2018}
A.~Athalye, N.~Carlini, and D.~Wagner.
\newblock Obfuscated gradients give a false sense of security: {Circumventing}
  defenses to adversarial examples.
\newblock \emph{Int. Conf. Mach. Learn.}, 2018.

\bibitem[Bradbury et~al.(2018)Bradbury, Frostig, Hawkins, Johnson, Leary,
  Maclaurin, and Wanderman-Milne]{bradbury_jax_2018}
J.~Bradbury, R.~Frostig, P.~Hawkins, M.~J. Johnson, C.~Leary, D.~Maclaurin, and
  S.~Wanderman-Milne.
\newblock {JAX}: composable transformations of {Python}+{NumPy} programs, 2018.
\newblock URL \url{http://github.com/google/jax}.

\bibitem[Carlini and Wagner(2017{\natexlab{a}})]{carlini_adversarial_2017}
N.~Carlini and D.~Wagner.
\newblock Adversarial examples are not easily detected: {Bypassing} ten
  detection methods.
\newblock In \emph{Proceedings of the 10th {ACM} {Workshop} on {Artificial}
  {Intelligence} and {Security}}, pages 3--14. ACM, 2017{\natexlab{a}}.

\bibitem[Carlini and Wagner(2017{\natexlab{b}})]{carlini_towards_2017}
N.~Carlini and D.~Wagner.
\newblock Towards evaluating the robustness of neural networks.
\newblock In \emph{2017 {IEEE} {Symposium} on {Security} and {Privacy}},
  2017{\natexlab{b}}.

\bibitem[Carmon et~al.(2019)Carmon, Raghunathan, Schmidt, Duchi, and
  Liang]{carmon_unlabeled_2019}
Y.~Carmon, A.~Raghunathan, L.~Schmidt, J.~C. Duchi, and P.~S. Liang.
\newblock Unlabeled data improves adversarial robustness.
\newblock In \emph{Adv. Neural Inform. Process. Syst.}, 2019.

\bibitem[Chen et~al.(2021)Chen, Zhang, Liu, Chang, and Wang]{chen2021robust}
T.~Chen, Z.~Zhang, S.~Liu, S.~Chang, and Z.~Wang.
\newblock Robust overfitting may be mitigated by properly learned smoothening.
\newblock In \emph{International Conference on Learning Representations},
  volume~1, 2021.

\bibitem[Croce and Hein(2020{\natexlab{a}})]{croce2020minimally}
F.~Croce and M.~Hein.
\newblock Minimally distorted adversarial examples with a fast adaptive
  boundary attack.
\newblock \emph{arXiv preprint arXiv:1907.02044}, 2020{\natexlab{a}}.
\newblock URL \url{https://arxiv.org/pdf/1907.02044}.

\bibitem[Croce and Hein(2020{\natexlab{b}})]{croce_reliable_2020}
F.~Croce and M.~Hein.
\newblock Reliable evaluation of adversarial robustness with an ensemble of
  diverse parameter-free attacks.
\newblock \emph{arXiv preprint arXiv:2003.01690}, 2020{\natexlab{b}}.

\bibitem[Croce et~al.(2020)Croce, Andriushchenko, Sehwag, Flammarion, Chiang,
  Mittal, and Hein]{croce2020robustbench}
F.~Croce, M.~Andriushchenko, V.~Sehwag, N.~Flammarion, M.~Chiang, P.~Mittal,
  and M.~Hein.
\newblock Robustbench: a standardized adversarial robustness benchmark.
\newblock \emph{arXiv preprint arXiv:2010.09670}, 2020.

\bibitem[Cubuk et~al.(2019)Cubuk, Zoph, Mane, Vasudevan, and
  Le]{cubuk_autoaugment:_2018}
E.~D. Cubuk, B.~Zoph, D.~Mane, V.~Vasudevan, and Q.~V. Le.
\newblock Autoaugment: {Learning} augmentation policies from data.
\newblock \emph{IEEE Conf. Comput. Vis. Pattern Recog.}, 2019.

\bibitem[Cubuk et~al.(2020)Cubuk, Zoph, Shlens, and Le]{cubuk2019randaugment}
E.~D. Cubuk, B.~Zoph, J.~Shlens, and Q.~V. Le.
\newblock Randaugment: Practical automated data augmentation with a reduced
  search space.
\newblock \emph{IEEE Conf. Comput. Vis. Pattern Recog.}, 2020.

\bibitem[Cui et~al.(2020)Cui, Liu, Wang, and Jia]{cui2020learnable}
J.~Cui, S.~Liu, L.~Wang, and J.~Jia.
\newblock Learnable boundary guided adversarial training.
\newblock \emph{arXiv preprint arXiv:2011.11164}, 2020.
\newblock URL \url{https://arxiv.org/pdf/2011.11164}.

\bibitem[DeVries and Taylor(2017)]{devries2017improved}
T.~DeVries and G.~W. Taylor.
\newblock Improved regularization of convolutional neural networks with cutout.
\newblock \emph{arXiv preprint arXiv:1708.04552}, 2017.

\bibitem[Dong et~al.(2018)Dong, Liao, Pang, Su, Zhu, Hu, and
  Li]{dong_boosting_2017}
Y.~Dong, F.~Liao, T.~Pang, H.~Su, J.~Zhu, X.~Hu, and J.~Li.
\newblock Boosting {Adversarial} {Attacks} with {Momentum}.
\newblock \emph{IEEE Conf. Comput. Vis. Pattern Recog.}, 2018.

\bibitem[Garipov et~al.(2018)Garipov, Izmailov, Podoprikhin, Vetrov, and
  Wilson]{garipov2018loss}
T.~Garipov, P.~Izmailov, D.~Podoprikhin, D.~Vetrov, and A.~G. Wilson.
\newblock Loss surfaces, mode connectivity, and fast ensembling of dnns.
\newblock \emph{arXiv preprint arXiv:1802.10026}, 2018.
\newblock URL \url{https://arxiv.org/pdf/1802.10026}.

\bibitem[Goodfellow et~al.(2015)Goodfellow, Shlens, and
  Szegedy]{goodfellow_explaining_2014}
I.~J. Goodfellow, J.~Shlens, and C.~Szegedy.
\newblock Explaining and harnessing adversarial examples.
\newblock \emph{Int. Conf. Learn. Represent.}, 2015.

\bibitem[Gowal et~al.(2019)Gowal, Uesato, Qin, Huang, Mann, and
  Kohli]{gowal_alternative_2019}
S.~Gowal, J.~Uesato, C.~Qin, P.-S. Huang, T.~Mann, and P.~Kohli.
\newblock An {Alternative} {Surrogate} {Loss} for {PGD}-based {Adversarial}
  {Testing}.
\newblock \emph{arXiv preprint arXiv:1910.09338}, 2019.

\bibitem[Gowal et~al.(2020)Gowal, Qin, Uesato, Mann, and
  Kohli]{gowal_uncovering_2020}
S.~Gowal, C.~Qin, J.~Uesato, T.~Mann, and P.~Kohli.
\newblock Uncovering the limits of adversarial training against norm-bounded
  adversarial examples.
\newblock \emph{arXiv preprint arXiv:2010.03593}, 2020.
\newblock URL \url{https://arxiv.org/pdf/2010.03593}.

\bibitem[Goyal et~al.(2017)Goyal, Dollár, Girshick, Noordhuis, Wesolowski,
  Kyrola, Tulloch, Jia, and He]{goyal2017accurate}
P.~Goyal, P.~Dollár, R.~Girshick, P.~Noordhuis, L.~Wesolowski, A.~Kyrola,
  A.~Tulloch, Y.~Jia, and K.~He.
\newblock Accurate, large minibatch sgd: Training imagenet in 1 hour.
\newblock \emph{arXiv preprint arXiv:1706.02677}, 2017.

\bibitem[Grefenstette et~al.(2018)Grefenstette, Stanforth, O'Donoghue, Uesato,
  Swirszcz, and Kohli]{grefenstette2018strength}
E.~Grefenstette, R.~Stanforth, B.~O'Donoghue, J.~Uesato, G.~Swirszcz, and
  P.~Kohli.
\newblock Strength in numbers: Trading-off robustness and computation via
  adversarially-trained ensembles.
\newblock \emph{arXiv preprint arXiv:1811.09300}, 2018.
\newblock URL \url{https://arxiv.org/pdf/1811.09300}.

\bibitem[He et~al.(2016)He, Zhang, Ren, and Sun]{he2015deep}
K.~He, X.~Zhang, S.~Ren, and J.~Sun.
\newblock Deep residual learning for image recognition.
\newblock \emph{IEEE Conf. Comput. Vis. Pattern Recog.}, 2016.

\bibitem[Hendrycks and Gimpel(2016)]{hendrycks2016gaussian}
D.~Hendrycks and K.~Gimpel.
\newblock Gaussian error linear units (gelus).
\newblock \emph{arXiv preprint arXiv:1606.08415}, 2016.

\bibitem[Hendrycks et~al.(2019)Hendrycks, Lee, and
  Mazeika]{hendrycks_using_2019}
D.~Hendrycks, K.~Lee, and M.~Mazeika.
\newblock Using {Pre}-{Training} {Can} {Improve} {Model} {Robustness} and
  {Uncertainty}.
\newblock \emph{Int. Conf. Mach. Learn.}, 2019.

\bibitem[Huang et~al.(2020)Huang, Zhang, and Zhang]{huang_self-adaptive_2020}
L.~Huang, C.~Zhang, and H.~Zhang.
\newblock Self-{Adaptive} {Training}: beyond {Empirical} {Risk} {Minimization}.
\newblock \emph{arXiv preprint arXiv:2002.10319}, 2020.

\bibitem[Izmailov et~al.(2018)Izmailov, Podoprikhin, Garipov, Vetrov, and
  Wilson]{izmailov_averaging_2018}
P.~Izmailov, D.~Podoprikhin, T.~Garipov, D.~Vetrov, and A.~G. Wilson.
\newblock Averaging {Weights} {Leads} to {Wider} {Optima} and {Better}
  {Generalization}.
\newblock \emph{Uncertainty in Artificial Intelligence}, 2018.

\bibitem[Kannan et~al.(2018)Kannan, Kurakin, and
  Goodfellow]{kannan_adversarial_2018}
H.~Kannan, A.~Kurakin, and I.~Goodfellow.
\newblock Adversarial {Logit} {Pairing}.
\newblock \emph{arXiv preprint arXiv:1803.06373}, 2018.

\bibitem[Kingma and Ba(2014)]{kingma_adam:_2014}
D.~P. Kingma and J.~Ba.
\newblock Adam: {A} method for stochastic optimization.
\newblock \emph{arXiv preprint arXiv:1412.6980}, 2014.

\bibitem[Krizhevsky et~al.(2009)Krizhevsky, Hinton,
  et~al.]{krizhevsky2009learning}
A.~Krizhevsky, G.~Hinton, et~al.
\newblock Learning multiple layers of features from tiny images.
\newblock 2009.

\bibitem[Kurakin et~al.(2016)Kurakin, Goodfellow, and
  Bengio]{kurakin_adversarial_2016}
A.~Kurakin, I.~Goodfellow, and S.~Bengio.
\newblock Adversarial examples in the physical world.
\newblock \emph{ICLR workshop}, 2016.

\bibitem[Lee et~al.(2020)Lee, Zaheer, Astrid, and Lee]{lee2020smoothmix}
J.-H. Lee, M.~Z. Zaheer, M.~Astrid, and S.-I. Lee.
\newblock Smoothmix: a simple yet effective data augmentation to train robust
  classifiers.
\newblock \emph{IEEE Conf. Comput. Vis. Pattern Recog. Worksh.}, 2020.

\bibitem[Madry et~al.(2018)Madry, Makelov, Schmidt, Tsipras, and
  Vladu]{madry_towards_2017}
A.~Madry, A.~Makelov, L.~Schmidt, D.~Tsipras, and A.~Vladu.
\newblock Towards deep learning models resistant to adversarial attacks.
\newblock \emph{Int. Conf. Learn. Represent.}, 2018.

\bibitem[Mosbach et~al.(2018)Mosbach, Andriushchenko, Trost, Hein, and
  Klakow]{mosbach_logit_2018}
M.~Mosbach, M.~Andriushchenko, T.~Trost, M.~Hein, and D.~Klakow.
\newblock Logit {Pairing} {Methods} {Can} {Fool} {Gradient}-{Based} {Attacks}.
\newblock \emph{arXiv preprint arXiv:1810.12042}, 2018.

\bibitem[Najafi et~al.(2019)Najafi, Maeda, Koyama, and
  Miyato]{najafi_robustness_2019}
A.~Najafi, S.-i. Maeda, M.~Koyama, and T.~Miyato.
\newblock Robustness to adversarial perturbations in learning from incomplete
  data.
\newblock \emph{Adv. Neural Inform. Process. Syst.}, 2019.

\bibitem[Nesterov(1983)]{nesterov27method}
Y.~Nesterov.
\newblock A method of solving a convex programming problem with convergence
  rate $o(1/k^{2})$.
\newblock In \emph{Sov. Math. Dokl}, 1983.

\bibitem[Netzer et~al.(2011)Netzer, Wang, Coates, Bissacco, Wu, and
  Ng]{netzer2011reading}
Y.~Netzer, T.~Wang, A.~Coates, A.~Bissacco, B.~Wu, and A.~Y. Ng.
\newblock Reading digits in natural images with unsupervised feature learning.
\newblock 2011.

\bibitem[Pang et~al.(2019)Pang, Xu, Du, Chen, and Zhu]{pang2019improving}
T.~Pang, K.~Xu, C.~Du, N.~Chen, and J.~Zhu.
\newblock Improving adversarial robustness via promoting ensemble diversity.
\newblock \emph{Int. Conf. Mach. Learn.}, 2019.

\bibitem[Pang et~al.(2020)Pang, Yang, Dong, Xu, Su, and
  Zhu]{pang_boosting_2020}
T.~Pang, X.~Yang, Y.~Dong, K.~Xu, H.~Su, and J.~Zhu.
\newblock Boosting {Adversarial} {Training} with {Hypersphere} {Embedding}.
\newblock \emph{Adv. Neural Inform. Process. Syst.}, 2020.

\bibitem[Papernot et~al.(2016)Papernot, McDaniel, Wu, Jha, and
  Swami]{papernot_distillation_2015}
N.~Papernot, P.~McDaniel, X.~Wu, S.~Jha, and A.~Swami.
\newblock Distillation as a defense to adversarial perturbations against deep
  neural networks.
\newblock \emph{IEEE Symposium on Security and Privacy}, 2016.

\bibitem[Polyak(1964)]{polyak1964some}
B.~T. Polyak.
\newblock Some methods of speeding up the convergence of iteration methods.
\newblock \emph{USSR Computational Mathematics and Mathematical Physics}, 1964.

\bibitem[Qin et~al.(2019)Qin, Martens, Gowal, Krishnan, Dvijotham, Fawzi, De,
  Stanforth, and Kohli]{qin_adversarial_2019}
C.~Qin, J.~Martens, S.~Gowal, D.~Krishnan, K.~Dvijotham, A.~Fawzi, S.~De,
  R.~Stanforth, and P.~Kohli.
\newblock Adversarial {Robustness} through {Local} {Linearization}.
\newblock \emph{Adv. Neural Inform. Process. Syst.}, 2019.

\bibitem[Rice et~al.(2020)Rice, Wong, and Kolter]{rice_overfitting_2020}
L.~Rice, E.~Wong, and J.~Z. Kolter.
\newblock Overfitting in adversarially robust deep learning.
\newblock \emph{Int. Conf. Mach. Learn.}, 2020.

\bibitem[Russakovsky et~al.(2015)Russakovsky, Deng, Su, Krause, Satheesh, Ma,
  Huang, Karpathy, Khosla, Bernstein, et~al.]{russakovsky2015imagenet}
O.~Russakovsky, J.~Deng, H.~Su, J.~Krause, S.~Satheesh, S.~Ma, Z.~Huang,
  A.~Karpathy, A.~Khosla, M.~Bernstein, et~al.
\newblock Imagenet large scale visual recognition challenge.
\newblock \emph{International journal of computer vision}, 2015.

\bibitem[Santurkar et~al.(2019)Santurkar, Tsipras, Tran, Ilyas, Engstrom, and
  Madry]{santurkar2019image}
S.~Santurkar, D.~Tsipras, B.~Tran, A.~Ilyas, L.~Engstrom, and A.~Madry.
\newblock Image synthesis with a single (robust) classifier.
\newblock \emph{arXiv preprint arXiv:1906.09453}, 2019.

\bibitem[Song et~al.(2019)Song, Shokri, and Mittal]{song2019privacy}
L.~Song, R.~Shokri, and P.~Mittal.
\newblock Privacy risks of securing machine learning models against adversarial
  examples.
\newblock In \emph{Proceedings of the 2019 ACM SIGSAC Conference on Computer
  and Communications Security}, 2019.

\bibitem[Strauss et~al.(2017)Strauss, Hanselmann, Junginger, and
  Ulmer]{strauss2017ensemble}
T.~Strauss, M.~Hanselmann, A.~Junginger, and H.~Ulmer.
\newblock Ensemble methods as a defense to adversarial perturbations against
  deep neural networks.
\newblock \emph{arXiv preprint arXiv:1709.03423}, 2017.

\bibitem[Szegedy et~al.(2014)Szegedy, Zaremba, Sutskever, Bruna, Erhan,
  Goodfellow, and Fergus]{szegedy_intriguing_2013}
C.~Szegedy, W.~Zaremba, I.~Sutskever, J.~Bruna, D.~Erhan, I.~Goodfellow, and
  R.~Fergus.
\newblock Intriguing properties of neural networks.
\newblock \emph{Int. Conf. Learn. Represent.}, 2014.

\bibitem[Takahashi et~al.(2018)Takahashi, Matsubara, and
  Uehara]{takahashi2018ricap}
R.~Takahashi, T.~Matsubara, and K.~Uehara.
\newblock Ricap: Random image cropping and patching data augmentation for deep
  cnns.
\newblock \emph{Asian Conf. Mach. Learn.}, 2018.

\bibitem[Torralba et~al.(2008)Torralba, Fergus, and Freeman]{80m}
A.~Torralba, R.~Fergus, and W.~T. Freeman.
\newblock 80 million tiny images: a large dataset for non-parametric object and
  scene recognition.
\newblock \emph{IEEE Trans. Pattern Anal. Mach. Intell.}, 2008.

\bibitem[Tramèr et~al.(2017)Tramèr, Kurakin, Papernot, Goodfellow, Boneh, and
  McDaniel]{tramer_ensemble_2017}
F.~Tramèr, A.~Kurakin, N.~Papernot, I.~Goodfellow, D.~Boneh, and P.~McDaniel.
\newblock Ensemble {Adversarial} {Training}: {Attacks} and {Defenses}.
\newblock \emph{arXiv preprint arXiv:1705.07204}, 2017.
\newblock URL \url{https://arxiv.org/pdf/1705.07204}.

\bibitem[Uesato et~al.(2018)Uesato, O'Donoghue, Oord, and
  Kohli]{uesato_adversarial_2018}
J.~Uesato, B.~O'Donoghue, A.~v.~d. Oord, and P.~Kohli.
\newblock Adversarial {Risk} and the {Dangers} of {Evaluating} {Against} {Weak}
  {Attacks}.
\newblock \emph{Int. Conf. Mach. Learn.}, 2018.

\bibitem[Uesato et~al.(2019)Uesato, Alayrac, Huang, Stanforth, Fawzi, and
  Kohli]{uesato_are_2019}
J.~Uesato, J.-B. Alayrac, P.-S. Huang, R.~Stanforth, A.~Fawzi, and P.~Kohli.
\newblock Are labels required for improving adversarial robustness?
\newblock \emph{Adv. Neural Inform. Process. Syst.}, 2019.

\bibitem[Wightman(2019)]{rw2019timm}
R.~Wightman.
\newblock Pytorch image models.
\newblock \url{https://github.com/rwightman/pytorch-image-models}, 2019.

\bibitem[Wu et~al.(2020)Wu, Xia, and Wang]{wu2020adversarial}
D.~Wu, S.-t. Xia, and Y.~Wang.
\newblock Adversarial weight perturbation helps robust generalization.
\newblock \emph{Adv. Neural Inform. Process. Syst.}, 2020.

\bibitem[Xie et~al.(2019)Xie, Wu, van~der Maaten, Yuille, and
  He]{xie_feature_2018}
C.~Xie, Y.~Wu, L.~van~der Maaten, A.~Yuille, and K.~He.
\newblock Feature denoising for improving adversarial robustness.
\newblock \emph{IEEE Conf. Comput. Vis. Pattern Recog.}, 2019.

\bibitem[Yun et~al.(2019)Yun, Han, Oh, Chun, Choe, and Yoo]{yun2019cutmix}
S.~Yun, D.~Han, S.~J. Oh, S.~Chun, J.~Choe, and Y.~Yoo.
\newblock Cutmix: Regularization strategy to train strong classifiers with
  localizable features.
\newblock \emph{Int. Conf. Comput. Vis.}, 2019.

\bibitem[Zagoruyko and Komodakis(2016)]{zagoruyko2016wide}
S.~Zagoruyko and N.~Komodakis.
\newblock Wide residual networks.
\newblock \emph{Brit. Mach. Vis. Conf.}, 2016.

\bibitem[Zhai et~al.(2019)Zhai, Cai, He, Dan, He, Hopcroft, and
  Wang]{zhai_adversarially_2019}
R.~Zhai, T.~Cai, D.~He, C.~Dan, K.~He, J.~Hopcroft, and L.~Wang.
\newblock Adversarially {Robust} {Generalization} {Just} {Requires} {More}
  {Unlabeled} {Data}.
\newblock \emph{arXiv preprint arXiv:1906.00555}, 2019.

\bibitem[Zhang et~al.(2017)Zhang, Bengio, Hardt, Recht, and
  Vinyals]{zhang2017understanding}
C.~Zhang, S.~Bengio, M.~Hardt, B.~Recht, and O.~Vinyals.
\newblock Understanding deep learning requires rethinking generalization.
\newblock \emph{Int. Conf. Learn. Represent.}, 2017.
\newblock URL \url{https://openreview.net/pdf?id=Sy8gdB9xx}.

\bibitem[Zhang et~al.(2018)Zhang, Cisse, Dauphin, and
  Lopez-Paz]{zhang2017mixup}
H.~Zhang, M.~Cisse, Y.~N. Dauphin, and D.~Lopez-Paz.
\newblock mixup: Beyond empirical risk minimization.
\newblock \emph{Int. Conf. Learn. Represent.}, 2018.

\bibitem[Zhang et~al.(2019)Zhang, Yu, Jiao, Xing, Ghaoui, and
  Jordan]{zhang_theoretically_2019}
H.~Zhang, Y.~Yu, J.~Jiao, E.~P. Xing, L.~E. Ghaoui, and M.~I. Jordan.
\newblock Theoretically {Principled} {Trade}-off between {Robustness} and
  {Accuracy}.
\newblock \emph{Int. Conf. Mach. Learn.}, 2019.

\end{thebibliography}
\bibliographystyle{abbrvnat}

\clearpage
\onecolumn
\appendix

\section{Analysis of models}

In this section, we perform additional diagnostics that give us confidence that our models are not doing any form of gradient obfuscation or masking \citep{athalye_obfuscated_2018,uesato_adversarial_2018}.

\paragraph{\autoattack and robustness against black-box attacks.}

First, we report in \autoref{table:autoattack} the robust accuracy obtained by our strongest models against a diverse set of attacks.
These attacks are run as a cascade using the \autoattack library available at \url{https://github.com/fra31/auto-attack}.
The cascade is composed as follows:
\squishlist
    \item \textsc{AutoPGD-ce}, an untargeted attack using \gls{pgd} with an adaptive step on the cross-entropy loss \citep{croce_reliable_2020},
    \item \textsc{AutoPGD-t}, a targeted attack using \gls{pgd} with an adaptive step on the difference of logits ratio \citep{croce_reliable_2020},
    \item \textsc{Fab-t}, a targeted attack which minimizes the norm of adversarial perturbations \citep{croce2020minimally},
    \item \textsc{Square}, a query-efficient black-box attack \citep{andriushchenko_square_2019}.
\squishend
First, we observe that our combination of attacks, denoted \textsc{AA+MT} matches the final robust accuracy measured by \autoattack.
Second, we also notice that the black-box attack (i.e., \textsc{Square}) does not find any additional adversarial examples.
Overall, these results indicate that our empirical measurement of robustness is meaningful and that our models do not obfuscate gradients.

\begin{table}[h]
\caption{Clean (without adversarial attacks) accuracy and robust accuracy (against the different stages of \autoattack) on \cifar obtained by different models with our method. We compare them with the results of~\cite{gowal_uncovering_2020} for a \wrn-70-16 in the case without additional data. Refer to \url{https://github.com/fra31/auto-attack} for more details.\label{table:autoattack}}
\begin{center}
\resizebox{1.\textwidth}{!}{
\begin{tabular}{l|cc|cccc|cc}
    \hline
    \cellcolor{header} \textsc{Model} & \cellcolor{header} \textsc{Norm} & \cellcolor{header} \textsc{Radius} & \cellcolor{header} \textsc{AutoPGD-ce} & \cellcolor{header} + \textsc{AutoPGD-t} & \cellcolor{header} + \textsc{Fab-t} & \cellcolor{header} + \textsc{Square} & \cellcolor{header} \textsc{Clean} & \cellcolor{header} \textsc{AA+MT} \TBstrut \\
    \hline
    \wrn-28-10 (CutMix) & \multirow{3}{*}{\linf} & \multirow{3}{*}{$\epsilon = 8/255$} & 61.01\% & 57.61\% & 57.61\% & 57.61\% & 86.22\% & 57.50\% \Tstrut \\
    \wrn-70-16 (CutMix) & & & 62.65\% & 60.07\% & 60.07\% & 60.07\% & 87.25\% & 60.07\%
    \\
    \wrn-70-16~\cite{gowal_uncovering_2020} & & & 59.39\% & 57.21\% & 57.20\% & 57.20\% & 85.29\% &  57.14\% \Bstrut\\
    \hline
\end{tabular}
}
\end{center}
\end{table}

\paragraph{Further analysis of gradient obfuscation.}

\begin{figure*}[h]
\centering
\subfigure[\label{fig:eps_sweep}]{\includegraphics[width=0.4\textwidth]{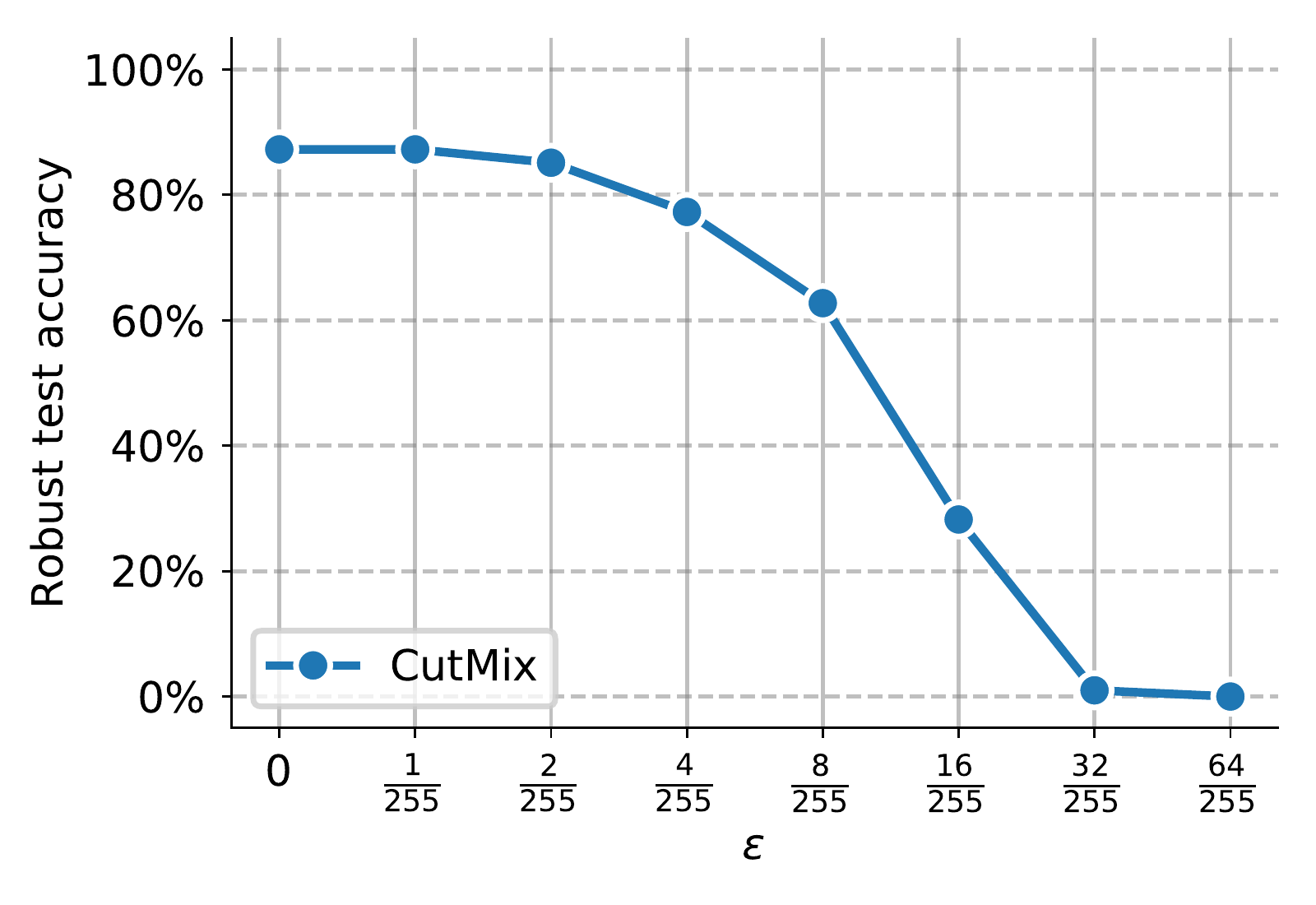}}
\subfigure[\label{fig:steps_sweep}]{\includegraphics[width=0.4\textwidth]{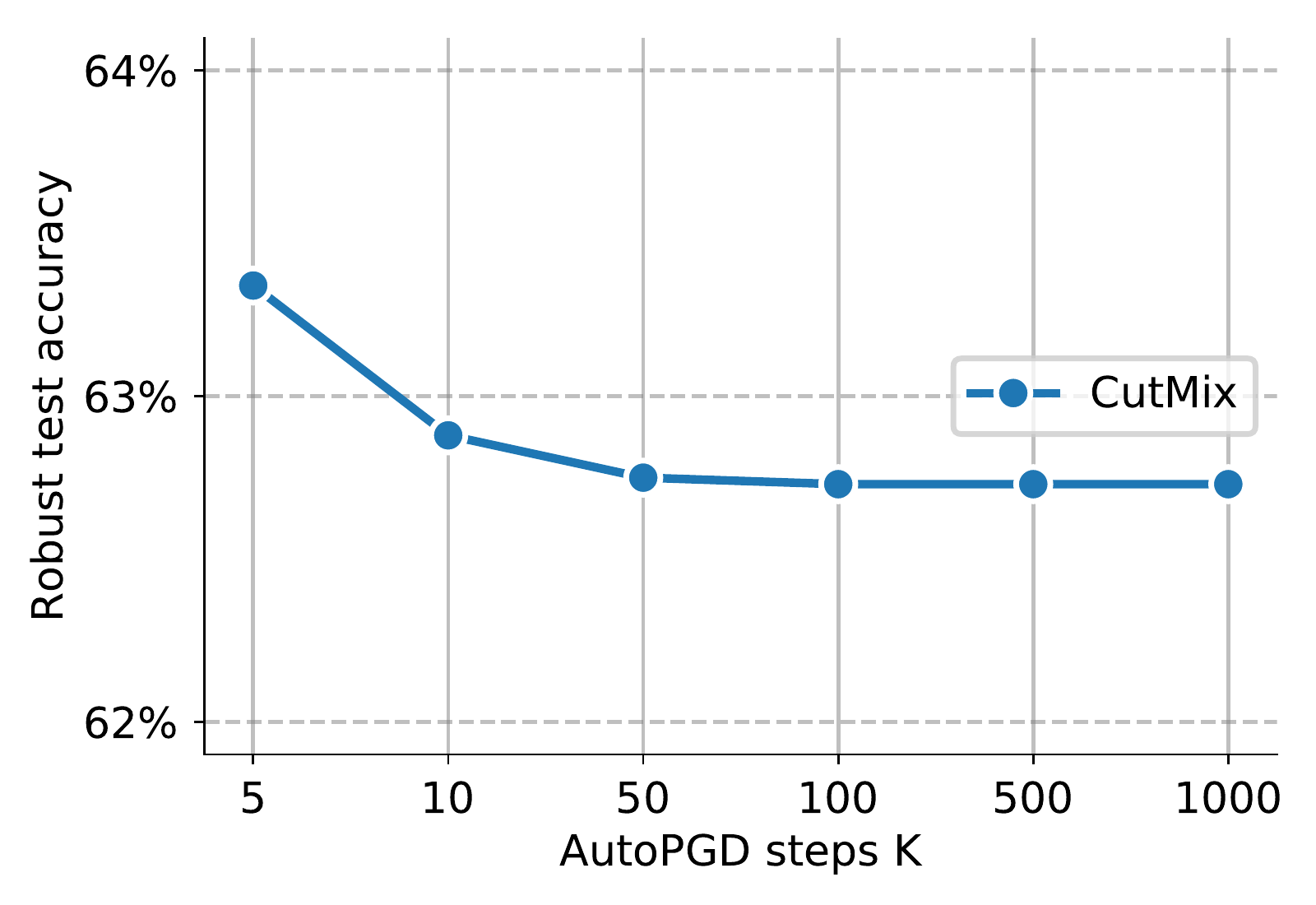}}
\caption{Robust test accuracy measured by running \textsc{AutoPGD-ce} with \subref{fig:eps_sweep} different radii $\epsilon_\infty$ and \subref{fig:steps_sweep} different number of steps $K$. The model is a \wrn-70-16 network trained with \emph{CutMix} against $\epsilon_\infty = 8/255$, which obtains 60.07\% robust accuracy against \textsc{AA+MT} at $\epsilon_\infty = 8/255$.}
\label{fig:grad_sweeps}
\end{figure*}

In this paragraph, we consider a \wrn-70-16 trained with \emph{CutMix} against $\epsilon_\infty = 8/255$, which obtains 60.07\% robust accuracy against \textsc{AA+MT} at $\epsilon_\infty = 8/255$.

In \autoref{fig:eps_sweep}, we run \textsc{AutoPGD-ce} with 100 steps and 1 restart and we vary the perturbation radius $\epsilon_\infty$ between zero and $64/255$.
As expected, the robust accuracy gradually drops as the radius increases indicating that \gls{pgd}-based attacks can find adversarial examples and are not hindered by gradient obfuscation.

In \autoref{fig:steps_sweep}, we run \textsc{AutoPGD-ce} with $\epsilon_\infty = 8/255$ and 1 restart and vary the number of steps $K$ between five and 1000.
We observe that the measured robust accuracy converges after 50 steps.
This is further indication that attacks converge in $100$ steps.

\paragraph{Loss landscapes.}

Finally, we analyze the adversarial loss landscapes of the model considered in the previous paragraph.
To generate a loss landscape, we vary the network input along the linear space defined by the worse perturbation found by \pgd{40} ($u$ direction) and a random Rademacher direction ($v$ direction).
The $u$ and $v$ axes represent the magnitude of the perturbation added in each of these directions respectively and the $z$ axis is the adversarial margin loss~\citep{carlini_towards_2017}: $z_y - \max_{i \neq y} z_i$ (i.e., a misclassification occurs when this value falls below zero).

\begin{figure*}[h]
\centering
\subfigure[Image of a horse]{\includegraphics[width=0.49\textwidth]{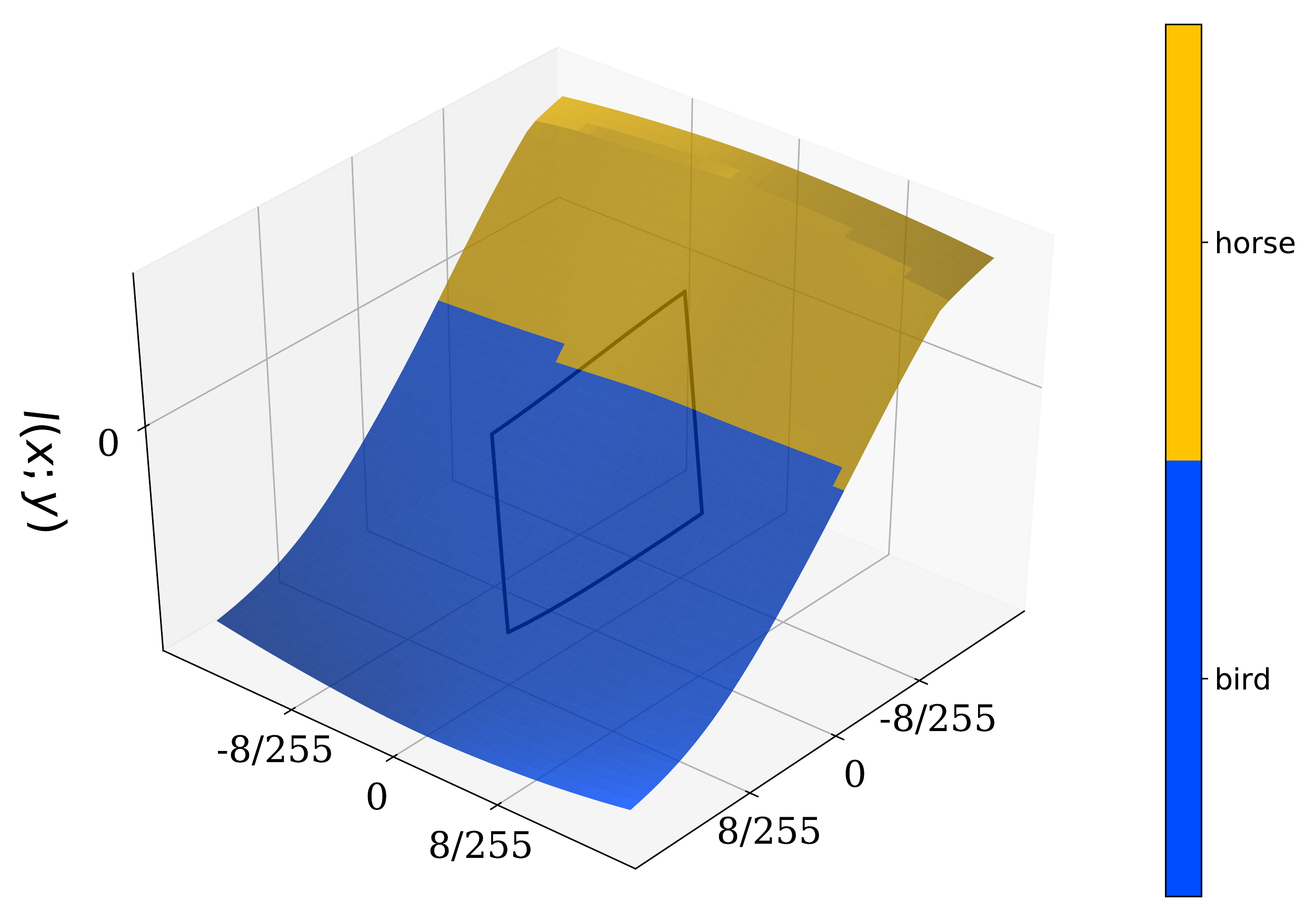}}
\subfigure[Image of an airplane]{\includegraphics[width=0.49\textwidth]{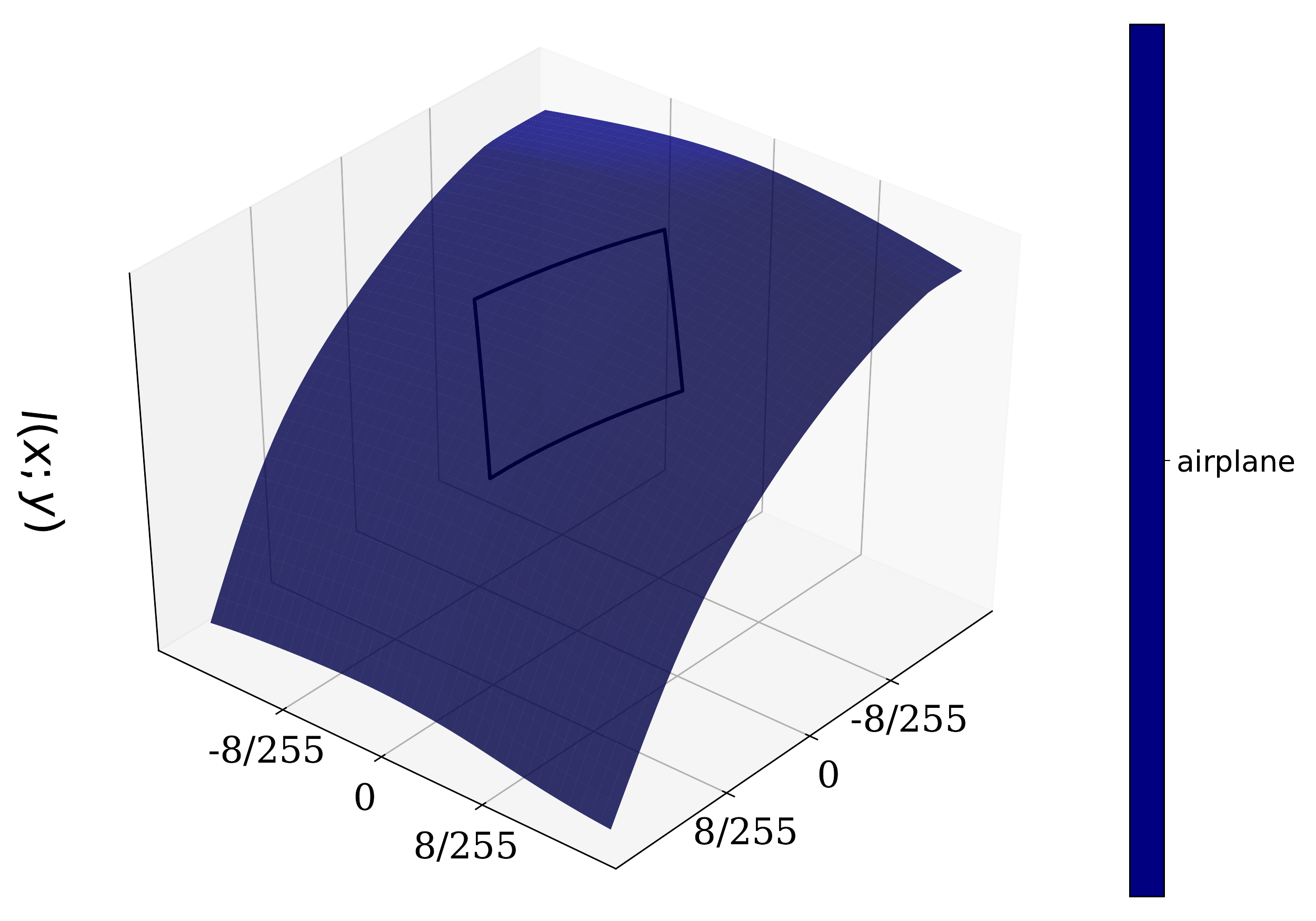}}
\caption{Loss landscapes around the first two images from the \cifar test set for the \wrn-70-16 network trained with \emph{CutMix}. This model obtains 60.07\% robust accuracy.
It is generated by varying the input to the model, starting from the original input image toward either the worst attack found using \pgd{40} ($u$ direction) or a random Rademacher direction ($v$ direction). The loss used for these plots is the margin loss $z_y - \max_{i \neq y} z_i$ (i.e., a misclassification occurs when this value falls below zero). The diamond-shape represents the projected \linf ball of size $\epsilon = 8/255$ around the nominal image.}
\label{fig:linf_landscapes}
\end{figure*}

\autoref{fig:linf_landscapes} shows the loss landscapes around the first 2 images of the \cifar test set.
All landscapes are smooth and do not exhibit patterns of gradient obfuscation.
Overall, it is difficult to interpret these figures further, but they do complement the numerical analyses done so far.

\section{Analysis of augmentations}
\label{app:aug}

In this section, we give the implementation details of the augmentations used in the paper and we complete our study with two additional augmentations: AutoAugment~\citep{cubuk_autoaugment:_2018} and RandAugment~\citep{cubuk2019randaugment}. Finally, we perform an ablation study analyzing the robust performance of each of the individual augmentations which compose RandAugment and we propose a curated version of RandAugment based on this ablation study.

\paragraph{Implementation details of the augmentations used in the paper.}

For \emph{Pad \& Crop}, we first pad the image by 4 pixels on each side and then take a random $32\times 32$ crop. For \emph{MixUp}~\citep{zhang2017mixup}, we sample the image mixing weight with a beta distribution $\operatorname{Beta}(\alpha, \alpha)$ with $\alpha=0.2$ in our default case. For \emph{Cutout}~\citep{devries2017improved}, we use a square window of size 16 whose center is uniformly sampled within the image bounds (but the window can overflow from the image). The square window is filled with the mean pixel values of the dataset. For \emph{CutMix}~\citep{yun2019cutmix}, we use a square window whose length is randomly sampled such that the patch area ratio follows a beta distribution $\operatorname{Beta}(\alpha, \beta)$ with parameters $\alpha = 1$ and $\beta = 1$ and whose center is uniformly sampled within the image bounds (but the window can overflow from the image). For \emph{RICAP}~\citep{takahashi2018ricap}, the size of the four patches is determined by the image center point whose coordinates are sampled with a beta distribution $\operatorname{Beta}(\alpha, \beta)$ with parameters $\alpha = 0.3$ and $\beta = 0.3$.

Regarding how data augmentation should be inserted in the adversarial training pipeline, there are two possible designs: applying the augmentation before or after the adversarial attack. Experimentally, we did find that applying the augmentation prior to running the attack was significantly better. Indeed, applying the augmentation after the attack significantly reduces the strength of the attack, since composition technique will destroy adversarial perturbations.

Regarding the training loss, we adapt the TRADES objective to the data augmentation setting leading us to minimize the following loss: $l_\textrm{ce}(f(\bf{x}'; \theta), y') + \beta D_\textrm{KL}(f(\bf{x}' + \delta'; \theta), f(\bf{x}'; \theta))$ where $\delta' = \textrm{argmax}_{\delta \in S} D_\textrm{KL}(f(\bf{x}' + \delta; \theta), y')$ with $\bf{x}'$ and $y'$ the result of various data augmentations. This formulation is inspired from \citet{gowal_uncovering_2020} where they explored various schemes for the inner optimization (see TRADES-XENT in \citep{gowal_uncovering_2020}).

\begin{table}[t]
 \caption{Delta in robust accuracy on the validation split of \cifar for each augmentation compared to the median robust accuracy over all the tested augmentations. The median robust accuracies are 53.61\% and 53.81\% for $M=3$ and $M=5$, respectively. We highlight in bold the results where the robust accuracy is significantly lower than the median robust accuracy.  \label{tab:ablation_aug}}%
\begin{center}
\resizebox{0.4\textwidth}{!}{
\begin{tabular}{l|rr}
    \hline

    \cellcolor{header} \textsc{Augmentations}  &  \cellcolor{header} M=3 & \cellcolor{header} M=5 \TBstrut \\
    \hline
    \hline
    AutoContrast & -0.20\% & 0.20\% \Tstrut \\         
    Equalize     & 0.78\% & 0.59\% \\          
    Invert       & \textbf{-6.84\%} & \textbf{-5.18\%} \\           
    Rotate       & 0.20\% & -0.39\% \\          
    Posterize    & -0.59\% & \textbf{-4.20\%} \\ 
    Solarize     & \textbf{-2.15\%} & 0.49\% \\  
    Color        & 0.39\% & 0.68\% \\ 
    Contrast     & 1.17\% & 0.20\% \\
    Brightness   & 0.98\% & 0.59\% \\  
    Sharpness    & -0.39\% & 0.10\% \\ 
    ShearX       & 0.00\% & -0.68\% \\
    ShearY       & -0.59\% & -0.10\% \\
    TranslateX   & 0.78\% & -0.39\% \\  
    TranslateY   & 0.10\% & 0.00\% \\   
    SolarizeAdd  & 0.00\% & -0.20\% \Bstrut \\  
    \hline
    \end{tabular}
}
\end{center}
\end{table}

\paragraph{Additional augmentations: AutoAugment and RandAugment.}

\emph{AutoAugment}~\citep{cubuk_autoaugment:_2018} learns augmentation policies from data by using a RNN controller and a Proximal
Policy Optimization algorithm. The learned policies available online have been tuned to optimize the natural accuracy and might not suit the adversarial setting. We evaluate in the adversarial setting these nominally trained policies by training a \wrn-28-10 on \cifar. By combining \emph{Pad \& Crop} and \emph{AutoAugment}, our model with WA reaches 55.68\% robust accuracy (and 86.51\% clean accuracy). Hence, the robust performance of \emph{AutoAugment} is similar to the one of \emph{MixUp} (i.e. 55.62\% robust accuracy) and it is performing much worse than \emph{CutMix} which gets 57.50\% robust accuracy.

\emph{RandAugment}~\citep{cubuk2019randaugment}, based on \emph{AutoAugment}, has similar performance in the standard training setting but it has a much smaller search space as it requires only two search sweeps: one over the length of the transformations sequence $N$ and a second over the global transformation magnitude $M$. This allows to quickly adapt \emph{RandAugment} to the adversarial setting compared to \emph{AutoAugment} and its computationally expensive search algorithm. We based our implementation of \emph{RandAugment} on the version found in the \emph{timm} library~\citep{rw2019timm} as in the original code of \emph{RandAugment} a small global magnitude $M$ can produce high magnitude transformations for some augmentations like \emph{Posterize}. After running a sweep over $N=\{1,2,3\}$ and $M=\{1,3,5,7\}$ when training adversarially a \wrn-28-10 on \cifar, the best model using \emph{RandAugment} with WA reaches 55.46\% robust accuracy against \textsc{AA+MT} (and 86.65\% clean accuracy), which a robust performance similar to the one obtained with \emph{AutoAugment}.

Finally, we noted in the main paper that, in the context of adversarial training, some augmentation techniques can perform worse than others such as \emph{MixUp} which leads to underfitting. So, we should study separately how each of the \emph{RandAugment} individual transformations performs for adversarial robustness. For each studied augmentation, we proceed as follows: (1) restrict the pool of available operations to the studied augmentation, (2) run RandAugment with $N=1$ and $M=\{3,5\}$ to train a \wrn-28-10 on \cifar. To compare the augmentations against each other, we evaluate the trained models on the validation split, and report in \autoref{tab:ablation_aug} the delta in robust accuracy compared to the median robust accuracy over all the tested transformations. 
We note that some augmentations, namely \emph{Invert}, \emph{Posterize} and \emph{Solarize}, significantly hurt robustness compared to the other augmentations. Hence, we run a curated version of RandAugment without these operations. After running a sweep over $N=\{1,2,3\}$ and $M=\{1,3,5,7\}$ when training adversarially a \wrn-28-10 on \cifar, the best model using this curated \emph{RandAugment} with WA reaches 56.51\% robust accuracy against \textsc{AA+MT} (and 86.30\% clean accuracy). Hence, the curated version of \emph{RandAugment} leads to an improvement of +1.05\% in robust accuracy compared to the non-curated version but is still worse than \emph{CutMix}. Nevertheless, \emph{RandAugment} is complementary to spatial composition techniques such as \emph{CutMix}. So, when combining the curated \emph{RandAugment} with \emph{CutMix} and re-running the same search sweeps, the best model with WA reaches 58.09\% robust accuracy against \textsc{AA+MT}, an improvement of +0.59\% over \emph{CutMix} alone with WA.

\section{Additional data setting}

For completeness, we also evaluate our models in the setting that considers additional external data extracted from \tinyimages~\cite{carmon_unlabeled_2019}.
In this setting (with 500K added images), we observe in Table~\ref{tab:combining_gen_aug_extra} that \emph{CutMix} performs worse than the baseline when the model is small (i.e., \wrn-28-10).
This is expected as the external data should generally be more useful than the augmented data and the model capacity is too low to take advantage of all this additional data.
However, \emph{CutMix} is beneficial in the setting with external data when the model is large (i.e., \wrn-70-16).
It improves upon the current state-of-the-art set by \citet{gowal_uncovering_2020} by +0.68\% (in the \linf setting) and +1.19\% (in the \ltwo setting) in robust accuracy, thus leading to a new state-of-the-art robust accuracy of respectively 66.56\% and 82.23\% when using external data.

\begin{table}[h]
 \caption{Clean (without adversarial attacks) accuracy and robust accuracy (against \textsc{AA+MT}) on \cifar as we both test against $\epsilon_\infty = 8/255$ and $\epsilon_2 = 128/255$ in the setting with added images from \tinyimages.\label{tab:combining_gen_aug_extra}}%
\begin{center}
\resizebox{0.6\textwidth}{!}{
\begin{tabular}{l|cc|cc}
    \hline
    \cellcolor{header} \textsc{Setup}  & \multicolumn{2}{c}{\cellcolor{header} \linf} & \multicolumn{2}{c}{\cellcolor{header} \ltwo}  \Tstrut \\
    \cellcolor{header} & \cellcolor{header} \textsc{Clean} & \cellcolor{header} \textsc{Robust} & \cellcolor{header} \textsc{Clean} & \cellcolor{header} \textsc{Robust} \Bstrut \\
    \hline
    \hline
    \multicolumn{5}{l}{\cellcolor{subheader} \textsc{\wrn-28-10} } \TBstrut \\
    \hline
    \citet{gowal_uncovering_2020} (trained by us)  &  89.42\% & \textbf{63.05\%} & 94.01\% & 80.08\% \Tstrut \\
    Ours (CutMix) &  89.90\% & 62.06\% & 94.96\% & \textbf{80.96\%} \Bstrut\\
    \hline
    \hline
    \multicolumn{5}{l}{\cellcolor{subheader} \textsc{\wrn-70-16}} \TBstrut \\
    \hline
    \citet{gowal_uncovering_2020} (trained by us)  &  90.51\% & 65.88\% & 94.19\% & 81.04\%  \Tstrut \\
    Ours (CutMix) &  92.23\% & \textbf{66.56\%} & 95.74\% & \textbf{82.23\%} \Bstrut\\
    \hline
    \end{tabular}
}
\end{center}
\end{table}

\section{Societal impact}

Improving robustness can have a negative impact for various reasons: (1) one can create stronger adversarial images by maximizing the error of the improved model as done in the image synthesis paper by~\citet{santurkar2019image}, (2)  the increased influence of the training points on the model when using adversarial training can reduce the sensitivity of the model and lead to larger biases~\cite{zhang_theoretically_2019}, (3) \citet{song2019privacy} show that improved robustness increases the success of privacy attacks and that robust models are more sensitive to membership inference attacks.

\clearpage

\end{document}